%% file: example.tex
\documentclass{article}
\usepackage{soul}

\usepackage[preprint]{corl_2026} % Uncomment for pre-prints (e.g., arxiv); This is like ``final'', but will remove the CORL footnote.
\usepackage{amsmath, amssymb}
\usepackage{float}
\usepackage{graphicx}
\usepackage{wrapfig}
\usepackage{algorithm}
\usepackage{algpseudocode}
\usepackage{enumitem}

\title{VICX: Generalizable Robot Manipulation via Video Generation and In-Context Operator Network}

\author{
  \begin{minipage}{\textwidth}\centering
  \textbf{Song Chen$^{1,*}$,\quad
  Linyan Xiang$^{1,*}$,\quad
  Ying Zhou$^{1}$,\quad
  Liu Yang$^{1,\dagger}$}\\[3pt]
  $^{1}$National University of Singapore\\[3pt]
  \texttt{song.chen@nus.edu.sg,\quad lxiang@nus.edu.sg,\quad yingzhou@u.nus.edu,\quad yangliu@nus.edu.sg}\\[4pt]
  {\small $^{*}$Equal contribution.\quad$^{\dagger}$Corresponding author.}
  \end{minipage}
}

\begin{document}
\maketitle

%===============================================================================

\begin{abstract}
Generalizable robot manipulation requires not only task-level reasoning over unseen scenes, but also reliable grounding of visual plans into embodiment-specific execution. To bridge this gap, we propose VICX (Video generation and In-Context eXecution), a decoupled closed-loop manipulation framework. In VICX, a frozen video generation model produces vision-language-conditioned high-level visual plans, while a Video-to-Trajectory In-Context Operator Network (V2T-ICON) serves as the task-agnostic interface that grounds these plans into executable robot-state trajectories. To improve execution generalization, V2T-ICON operates on segmentation-extracted arm-only frame observations and uses retrieved image-state pairs as in-context prompts, allowing a robust and generalizable visual-to-state mapping at inference time without parameter updates. Experiments on Meta-World show that VICX supports cross-task generalization, closed-loop self-correction, and cross-embodiment transfer, demonstrating dual generalization across both task semantics and robot execution. The project webpage can be found here: \url{https://scaling-group.github.io/vicx/}.

% V3:
%Generalizable manipulation requires both scene-level reasoning and cross-embodiment execution. However, existing models often couple these capabilities inside a single robot-data-driven end-to-end policy, making generalization highly dependent on scarce embodiment-specific trajectories. We propose a decoupled closed-loop manipulation framework in which a frozen video generation model produces language-conditioned high-level visual plans, and Video-to-Trajectory In-Context Operator Network (V2T-ICON) serves as the task-agnostic execution interface that grounds these plans into executable robot-state trajectories. To improve execution generalization, V2T-ICON operates on segmentation-extracted arm-only frames and retrieves visually similar image-state trajectory segments as in-context prompts, allowing the video-to-trajectory operator to instantiate a query-specific visual-to-state mapping at inference time without parameter updates. Experiments on Meta-World show that this hierarchical design supports cross-task generalization, closed-loop self-correction, and cross-embodiment transfer, enabling a single framework to generalize simultaneously over what task to perform and which robot performs it. Our code and data are available at: https://scaling-group.github.io/vig-icon/

\end{abstract}

% Two or three meaningful keywords should be added here

\section{Introduction}\label{sec: intro}

\input{sections/introduction}
%===============================================================================

\section{Method}\label{sec: method}
\input{sections/method}

\section{Experimental Results}
\label{sec:result}

\input{sections/results}
\input{sections/cross_embodiment}

\input{sections/benefits_ICL}

%===============================================================================

\section{Related Work}
\label{sec:related_work}
\input{sections/related_work}

\section{Conclusion}
\label{sec:conclusion}
\input{sections/conclusion.tex}

\section{Limitations}
\label{sec:limitations}
\input{sections/limitation}

%===============================================================================

\clearpage
% The acknowledgments are automatically included only in the final and preprint versions of the paper.
% TODO(arxiv): Fill in acknowledgments before upload --- grant/funding numbers,
%   GPU/compute sponsorship (important for compute-heavy foundation-model work),
%   and people to thank. Placeholder text below.
\acknowledgments{Liu Yang acknowledges support from the National Research Foundation, Singapore, under the NRF fellowship (Project No. NRF-NRFF17-2025-0006). We acknowledge NUS IT’s Research Computing group for providing computational support.}

%===============================================================================

% no \bibliographystyle is required, since the corl style is automatically used.
\bibliography{example}  % .bib

%===============================================================================

\clearpage
\appendix
\section*{Appendix}

\section{Experimental Training Details}
\label{app:exp_train_details}

\input{sections/appendix_exp_train_details}

\section{Experimental Evaluation Details}
\label{app:exp_eval_details}

\input{sections/appendix_exp_eval_details}

\end{document}

%% file: sections/introduction.tex
Current Vision-Language-Action (VLA) models provide a compelling paradigm for robot manipulation by unifying visual perception, language understanding, and action prediction within a single end-to-end policy~\citep{zitkovich2023rt2, kim2025openvla, black2025pi, ma2026surveyofvla}. However, their success still depends heavily on large-scale robot-specific demonstrations, since the same model must learn both high-level task reasoning and low-level physical control from embodied trajectories. 
Despite major community efforts such as Open X-Embodiment~\citep{oneill2024openx}, high-quality robot data remains scarce, expensive, and difficult to scale across diverse embodiments, environments, and manipulation skills. 
Recent World-Action Models (WAMs)~\citep{wang2026world} recognize this bottleneck and explore a complementary direction: using pre-trained video generation backbones as physical priors, then post-training them with robot data to jointly model future visual states and executable actions~\citep{ye2026world}. 
This line of work suggests that video priors can provide much of the physical knowledge missing from scarce robot trajectories. 
However, this video-action alignment reduces the dependence on purely reactive observation-to-action learning, raising a core open question: how should predicted visual futures be grounded into executable robot control?

In this work, we focus on this vision-to-execution bridge. 
We leverage frozen video generation models as out-of-the-box ``intelligence engines'' for high-level pixel-space planning: given the current observation and a language instruction, a frozen video model predicts how the scene and robot arm would progress. 
The generated video, however, is only a visual plan and cannot be directly executed. 
For this execution component, we introduce a novel method inspired by In-Context Operator Network (ICON)~\citep{yang2023context}: a task-agnostic Video-to-Trajectory ICON (V2T-ICON),  which maps a generated video plan to an executable robot-state trajectory. Crucially, instead of memorizing a rigid, global pixel-to-state mapping, V2T-ICON utilizes retrieved image-state pairs as in-context prompts to locally calibrate the visual-to-state correspondence. This design allows the model to dynamically infer novel arm trajectories induced by unseen tasks from the provided context at test time without parameter updates. 
Unlike many video-planning pipelines that recover actions through inverse dynamics between adjacent frames~\cite{du2023learning} or  extract actions by geometrically computing motion from structured representations, such as optical flow~\cite{ko2024learning}, V2T-ICON predicts state trajectories directly, using states as a more geometric and controller-agnostic interface between visual planning and physical control. 
Moreover, by focusing solely on robot-state inference from arm-only visual observations, V2T-ICON remains task-agnostic: it estimates how the robot arm moves rather than what task is being performed.

To unlock generalizable robot manipulation with video generation models and V2T-ICON, we propose VICX, which integrates both modules into a closed-loop manipulation framework. The video generation model provides planning generalization by predicting how the scene and the robot itself could progress under instructions. V2T-ICON provides execution generalization by translating the planned frames into the robot's state space. With such planning-level and execution-level generalization, VICX exhibits remarkable generalization in terms of task, scene, and even robot embodiment, without any fine-tuning. Furthermore, to ensure long-term robustness, we implement the VICX framework as a closed-loop system, as shown in Fig.~\ref{fig: closed_loop_evaluation}. By continuously updating the video model with real-time observations, VICX grounds the visual intent in the physical environment, effectively mitigating compounding errors from generative distortions, translation drift, and external disturbances \cite{ye2026world}.

% By shifting the burden of intelligence to internet-scale video models and maintaining execution reliability through closed-loop task-agnostic translation, our system successfully bypasses the data bottlenecks of traditional VLA and WAV architectures. 
In summary, our core contributions are threefold:
\begin{itemize}[leftmargin=*]
    \item We propose VICX, a decoupled, closed-loop manipulation framework that uses a frozen video generation model for high-level visual planning and formulates the remaining execution problem as video-to-trajectory grounding.
    \item We introduce V2T-ICON, a task-agnostic in-context operator network that predicts robot-state trajectories from generated videos using retrieved image-state prompts, enabling execution generalization across unseen task-induced trajectories and shifted robot embodiments.
    \item We demonstrate strong generalization in Meta-World experiments: with V2T-ICON trained on only three source tasks, VICX generalizes across nine manipulation tasks, exhibits closed-loop self-correction, and transfers to a shifted robot embodiment without retraining.
\end{itemize}

%% file: sections/method.tex
VICX separates generalizable manipulation into two stages: high-level visual planning and low-level state execution. First, a frozen video generation model predicts a future execution video. Second, V2T-ICON grounds this visual plan into an executable robot-state trajectory. We elaborate on the framework protocol, V2T-ICON formulation, and network details below.

\subsection{VICX: Closed-loop Generalizable Manipulation Framework}

\begin{wrapfigure}{r}{0.55\linewidth}
      \centering
      \includegraphics[width=\linewidth]{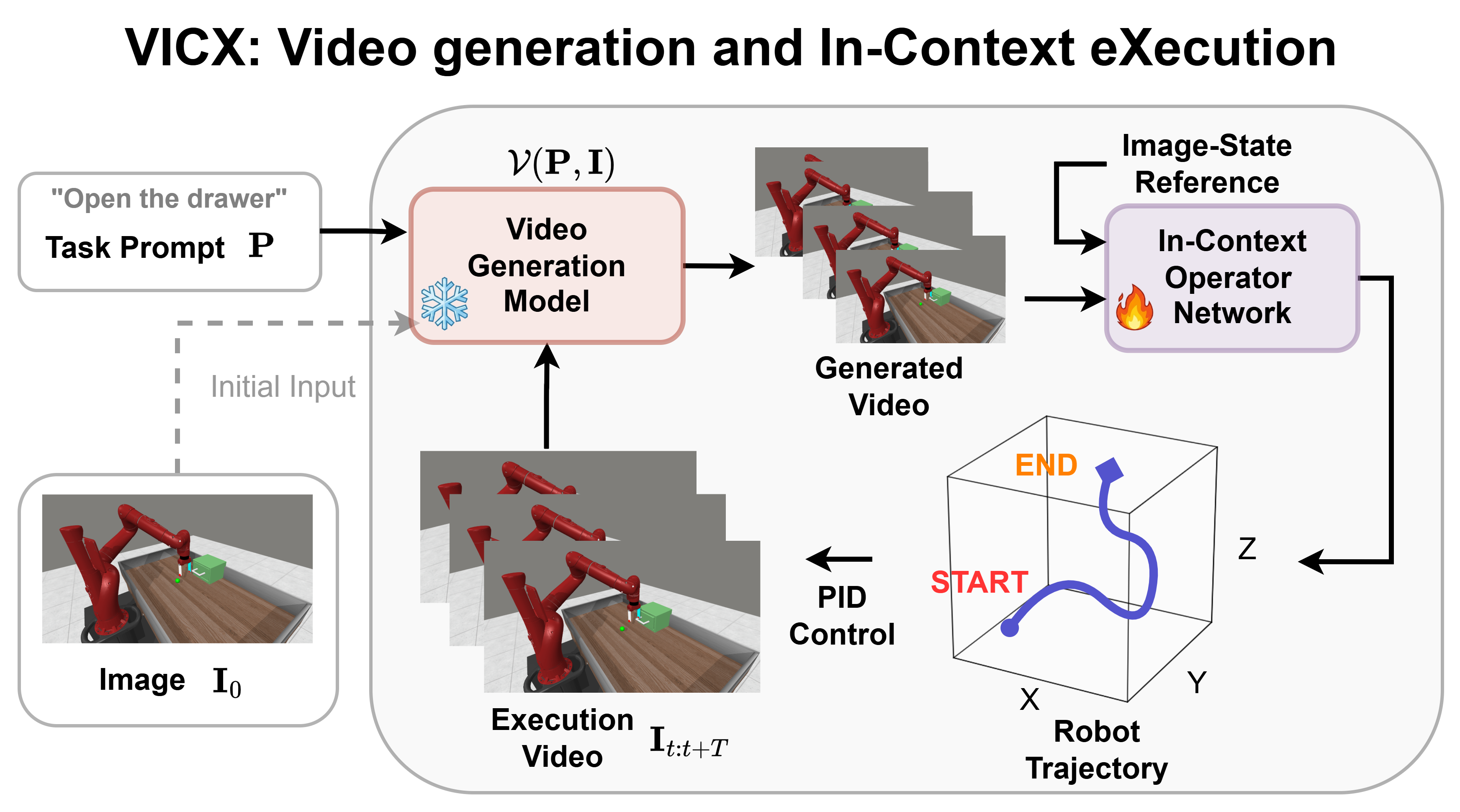}
      \caption{Overview of VICX, a closed-loop generalizable manipulation framework via video generation and video-to-trajectory in-context operator network (V2T-ICON). Given a prompt $P$ and current
  observation $I$, a frozen video generation model generates a visual plan, which
  is translated into a robot-state trajectory by V2T-ICON and executed with PID
  control. The resulting observations are fed back for the next replanning
  cycle.}
      \label{fig: closed_loop_evaluation}
  \end{wrapfigure}

As shown in Fig.~\ref{fig: closed_loop_evaluation}, a frozen video generation model can be directly used as a visual planner for manipulation without task-specific fine-tuning. Given the current observation $I_0$ and a task prompt $P$, the model predicts a short-horizon future video $\tilde{Q} = \mathcal{V}(P, I_0)$, which serves as a visual plan describing how the scene and the robot arm should evolve if the task is executed successfully. Importantly, this interface is model-agnostic: any image-conditioned video model, such as Wan, Sora, or Veo, can in principle be plugged into the same interface and used out of the box as a high-level planner.
The generated video itself cannot be directly executed by the robot. Therefore, after each visual plan is produced, V2T-ICON converts the planned frames into a sequence of robot states. These states are then tracked by a feedback controller in the environment. After execution, the last execution video is fed back to the video agent, which allows VICX to replan and correct accumulated errors. In this way, the video model provides high-level visual intent, while V2T-ICON grounds that intent in the robot's state and control space. The full manipulation pseudocode is provided in Appendix~\ref{app:video_agent_pseudocode}.

\subsection{Video-to-Trajectory In-Context Operator Network}

We address the problem of translating a single-view video of a robotic arm into its underlying state trajectory. A straightforward solution is to train a visual regressor that maps each frame independently to a robot state~\citep{tian2024robokeygen,labbe2021single,lee2020camera,ausserlechner2024zs6d,jantos2023poet}. However, it requires the network to learn a global mapping from pixels to states. Such a mapping is difficult to generalize because the same robot state can look different under different viewpoints, backgrounds, lighting conditions, robot appearances, and video-generation artifacts.

% \noindent \textbf{Video-To-Trajectory Translation as Operator Learning} We instead cast video-to-trajectory translation as an \emph{operator learning} problem. We first define the underlying video-to-trajectory operator in function space. Let $\Omega_{\mathrm{img}}$ denote the image domain and let $\mathcal{I} \subset L^2(\Omega_{\mathrm{img}};\mathbb{R}^{C})$ be the image function space, where $C$ is the number of image channels. A video can be viewed as a temporal image function $q \in \mathcal{Q} \subset L^2([0,T];\mathcal{I})$, while the corresponding robot trajectory is a state function $x \in \mathcal{X} \subset H^1([0,T];\mathbb{R}^{d_s})$, where $H^1$ reflects the temporal smoothness of robot motion. The learning object is therefore a video-to-trajectory operator
% $$
% \mathcal{G}: \mathcal{Q} \rightarrow \mathcal{X}, \qquad x = \mathcal{G}(q).
% $$
% In practice, we observe this operator through discrete samples at $T$ time steps. We write the sampled query video window as $Q = \{I_t^q\}_{t=1}^{T}$, where $I_t^q = q(\tau_t)$, and the corresponding sampled trajectory as $X = \{x_t\}_{t=1}^{T}$, where $x_t = x(\tau_t) \in \mathbb{R}^{d_s}$.
V2T-ICON takes a different approach. Instead of predicting the state from the query frame alone, it first retrieves visually similar image-state pairs from the training set and uses them as references. These references act like calibration examples. They tell the model, for the current visual domain, what robot states correspond to nearby arm appearances. The model then compares the query frames with these references and predicts a temporally consistent trajectory.

\noindent \textbf{Video-to-Trajectory In-Context Operator Network.} ICON learns mappings between function spaces by conditioning the operator on data prompts, enabling zero-shot generalization to new systems without parameter updates~\citep{yang2023context}. 
We cast video-to-trajectory translation as an ICON problem, where the operator maps a visual space to a robot-state trajectory space, i.e., $\mathcal{G}: Q \mapsto X$. 
Let $Q=\{I_t^q\}_{t=1}^{T}$ denote a query video window and $X=\{x_t\}_{t=1}^{T}$ the corresponding robot-state trajectory, with $x_t\in\mathbb{R}^{d_s}$. 
A standard visual regressor would learn a global mapping $X=\mathcal{G}(Q)$, but such a mapping is difficult to generalize across viewpoints, backgrounds, robot appearances, and video-generation artifacts. We therefore instantiate V2T-ICON as an in-context operator network~\citep{yang2023context}, where the mapping is conditioned on retrieved image-state references rather than represented only by fixed model parameters. For each query frame $I_t^q$, we retrieve $N$ visually similar reference images and their paired robot states, $R_t=\{(I_{t,j}^r,x_{t,j}^r)\}_{j=1}^{N}$. The full context for the video window is $R=\{R_t\}_{t=1}^{T}$, and V2T-ICON predicts the trajectory as
\[
\hat{X}=\mathcal{G}_{\theta}(Q,R)=\{\hat{x}_t\}_{t=1}^{T}.
\]
Here, $Q$ specifies the motion to be translated, while $R$ provides local calibration examples of how arm appearance maps to robot state in the current visual domain. This distinguishes V2T-ICON from a frame-wise visual regressor: instead of memorizing a single global pixel-to-state mapping, the model infers the query trajectory by comparing generated frames with retrieved visual-state examples. As a result, when a new generated video is observed at test time, V2T-ICON can adapt its translation through the retrieved context without updating its weights.

\begin{figure}[!t]
    \centering
    \includegraphics[width=1\linewidth]{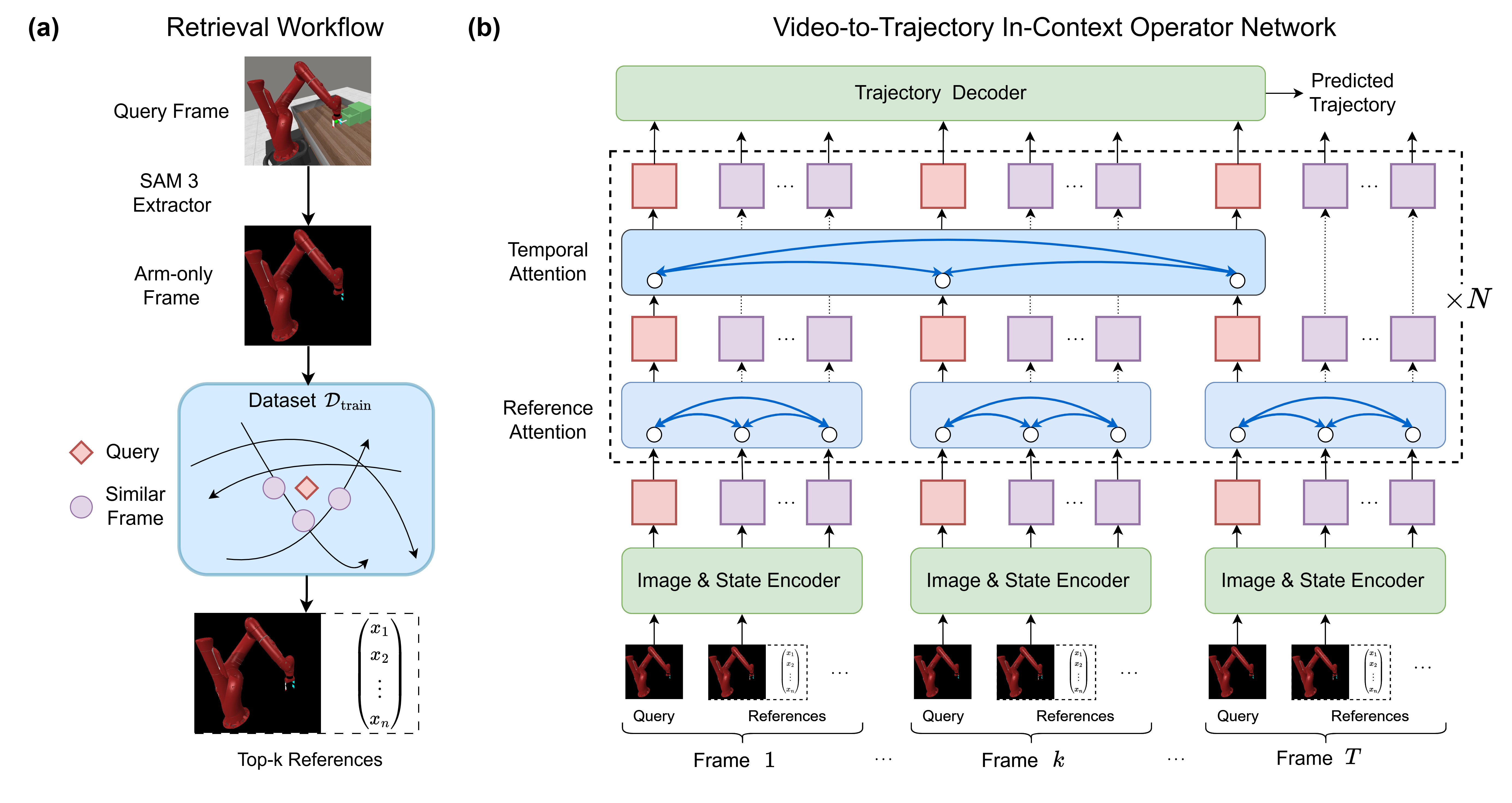}
    \caption{Overview of the Video-to-Trajectory In-Context Operator Network pipeline. \textbf{(a)} Reference retrieval: for each arm-only query frame, we search the arm-only training pool $\mathcal{D}_{\text{train}}$ and select the top-$N$ visually similar image-state pairs as references. \textbf{(b)} Operator network architecture: query and reference inputs are encoded by image and state encoders, reference attention performs per-frame in-context matching, temporal attention propagates information across frames, and the trajectory decoder predicts $\hat{X}$.}
    \label{fig: ICL_traj_trans}
\end{figure}

\noindent \textbf{Data Preparation.} Raw robot videos contain substantial task-specific scene content, which can introduce spurious correlations between state estimation and manipulation context. To reduce this bias and more accurately recover the arm trajectory from query videos, we apply SAM3-based extraction and retain only the robotic-arm body in each frame. Letting $\mathcal{S}(\cdot)$ denote the segmentation operator, each raw episode is transformed into
$E_{\mathrm{arm}}^{(m)} =
\{(\mathcal{S}(I_\ell^{(m)}), x_\ell^{(m)})\}_{\ell=1}^{L_m}.$
The final training pool is
$\mathcal{D}_{\text{train}} = \{E_{\mathrm{arm}}^{(m)}\}_{m=1}^{M}.$
Both query windows and retrieved references are sampled from this arm-only pool, encouraging V2T-ICON to rely on robot geometry, pose, and motion rather than objects or backgrounds. Detailed procedures for collecting the raw robot videos are provided in Appendix~\ref{app:tasks_dataset}.

\noindent \textbf{Reference Retrieval.} For each query frame, we use the arm-only image $\bar{I}_t^q=\mathcal{S}(I_t^q)$ for retrieval. Specifically, we search the arm-only training pool $\mathcal{D}_{\text{train}}$ by visual similarity and select the top-$N$ image-state pairs as references:
$R_t = \operatorname{TopN}_{(\bar{I}^r,x^r)\in\mathcal{D}_{\text{train}}}
\operatorname{sim}(\bar{I}_t^q,\bar{I}^r)
= \{(\bar{I}_{t,j}^r,x_{t,j}^r)\}_{j=1}^{N}.$
In implementation, similarity is computed using frozen DINOv2 features, and nearest-neighbor search is implemented with FAISS~\citep{douze2026faiss}. The retrieved references $R=\{R_t\}_{t=1}^{T}$ serve as the in-context prompt for V2T-ICON, as shown in Fig.~\ref{fig: ICL_traj_trans}(a).

\noindent \textbf{Training Objective.} During training, we sample a query window $(Q, X)$ from $\mathcal{D}_{\text{train}}$ and retrieve its reference context $R$. The model is trained to predict the ground-truth trajectory from the query video and its references. We use a weighted objective with three terms:
$
\min_{\theta}
\mathbb{E}_{(Q, X) \sim \mathcal{D}_{\text{train}}, \; R \sim p(R \mid Q, \mathcal{D}_{\text{train}})}
\left[
\lambda_{\mathrm{state}} \mathcal{L}_{\mathrm{state}}
\;+
\lambda_{\mathrm{smooth}} \mathcal{L}_{\mathrm{smooth}}
\;+
\lambda_{\mathrm{vel}} \mathcal{L}_{\mathrm{vel}}
\right].
$
The state loss supervises the predicted states, the smoothness loss discourages unrealistic jitter, and the velocity loss encourages the predicted motion trend to match the ground truth. The full mathematical definitions of these losses and their weights are given in Appendix~\ref{app:v2t_icon_training}.

\subsection{Architecture of the Video-to-Trajectory In-Context Operator Network}

Fig.~\ref{fig: ICL_traj_trans}(b) illustrates the V2T-ICON architecture. A simple design would concatenate all query and reference tokens over the full video window and process them with one Transformer. This is computationally expensive and mixes two distinct problems: matching each query frame to its references, and enforcing temporal consistency across frames. We therefore factorize the operator network into two attention stages: \emph{reference attention} performs per-frame in-context matching, and \emph{temporal attention} aggregates the resulting frame latents before the final trajectory decoder predicts $\hat{X}$. The detailed forward pseudocode of the V2T-ICON is provided in Appendix~\ref{app:v2t_icon_training}.

\noindent \textbf{Reference Attention.} Reference attention performs per-frame in-context matching. For each query frame, it compares the query arm appearance with the retrieved image-state references and produces a frame latent grounded by nearby state examples. This stage is designed to make state estimation depend on local visual correspondence rather than on a single global pixel-to-state regressor.

\noindent \textbf{Temporal Attention.} Temporal attention propagates information across the frame latents produced by reference attention. Its role is to resolve single-frame ambiguity, suppress noisy frame-wise estimates, and encourage physically smooth state evolution across the video window. This separation lets V2T-ICON first infer local visual-state correspondence and then enforce trajectory-level consistency.

\noindent \textbf{Iterative Refinement.} V2T-ICON repeats the reference-attention and temporal-attention stages for multiple refinement rounds. Early rounds obtain a coarse translation from retrieved examples, while later rounds use the previous temporally contextualized representation as a query-side prior.  This iterative design allows the model to first obtain a rough state estimate from retrieved references and then improve it using temporal context. As a result, V2T-ICON can translate generated visual plans into smoother and more executable robot trajectories.

%% file: sections/results.tex
We evaluate VICX in closed-loop simulation experiments on Meta-World manipulation tasks \cite{mclean2025metaworld}. The V2T-ICON is trained on only three source tasks, \texttt{drawer-open}, \texttt{reach}, and \texttt{basketball}. Additional details about the training datasets and model architecture are provided in Appendix~\ref{app:exp_train_details}. At test time, a frozen Wan video model proposes visual planning, and the V2T-ICON converts each plan into executable state trajectories using five retrieved image-state references per query frame. Unless otherwise stated, we use success rate as the primary performance metric, as defined in Appendix~\ref{app:evaluation_metrics}. The evaluation settings for all three baselines and VICX are described in Appendix~\ref{app:evaluation_results}. 

\subsection{Simulation Evaluation}
\label{sec:result:sim}

\begin{table}[h]
    \centering
    \small
    \caption{Closed-loop simulation success rates on the 9 Meta-World tasks, comparing the $\pi_{0.5}$-Scratch \cite{black2025pi}, $\pi_{0.5}$-Finetune, AVDC \cite{ko2024learning}, and VICX in the 5-Context setting. Each task is evaluated independently over 20 random seeds. Bold denotes the highest success rate for each task.}
    \begin{tabular}{lcccc}
        \hline
        Task & $\pi_{0.5}$-Scratch & $\pi_{0.5}$-Finetune & AVDC & VICX  \\
        \hline
        \texttt{button-press} & 5.0\% & 45.0\% & \textbf{60.0\%} & 20.0\% \\
        \texttt{button-press-topdown} & 0.0\% & 0.0\% & 60.0\% & \textbf{85.0\%} \\
        \texttt{coffee-button} & 0.0\% & 15.0\% & 50.0\% & \textbf{100.0\%} \\
        \texttt{door-close} & 0.0\% & 20.0\% & 85.0\% & \textbf{100.0\%} \\
        \texttt{drawer-close} & 95.0\% & 75.0\% & 65.0\% & \textbf{100.0\%} \\
        \texttt{drawer-open} & 0.0\% & 0.0\% & 20.0\% & \textbf{75.0\%} \\
        \texttt{faucet-close} & 10.0\% & 5.0\% & \textbf{50.0\%} & 25.0\% \\
        \texttt{faucet-open} & 5.0\% & 20.0\% & 25.0\% & \textbf{45.0\%} \\
        \texttt{handle-press} & 70.0\% & 55.0\% & 80.0\% & \textbf{100.0\%} \\
        \hline
        Overall & 20.6\% & 26.1\% & 55.0\% & \textbf{72.2\%} \\
        \hline
    \end{tabular}
    \label{tab:sim_main_results}
\end{table}

\noindent \textbf{Evaluation on Meta-World Tasks.}
Table~\ref{tab:sim_main_results} reports the closed-loop performance of VICX and three baseline policies on the same nine-task suite. Across the evaluated tasks, VICX achieves a 72.2\% overall success rate, outperforming $\pi_{0.5}$-Scratch, $\pi_{0.5}$-Finetune, and AVDC. This comparison is especially notable because $\pi_{0.5}$-Finetune is fine-tuned on Meta-World MT50 and AVDC is trained on 11 Meta-World tasks, whereas the V2T-ICON component in VICX is trained using data collected from only three Meta-World tasks. Despite this narrower training support, our method obtains the highest overall success rate across the nine evaluation tasks and can generalize to unseen tasks. We attribute this strong generalization to the decoupled architecture and world priors, as detailed in Section~\ref{sec:result_generalization}.

\begin{figure}[h]
    \centering
    \includegraphics[width=\linewidth]{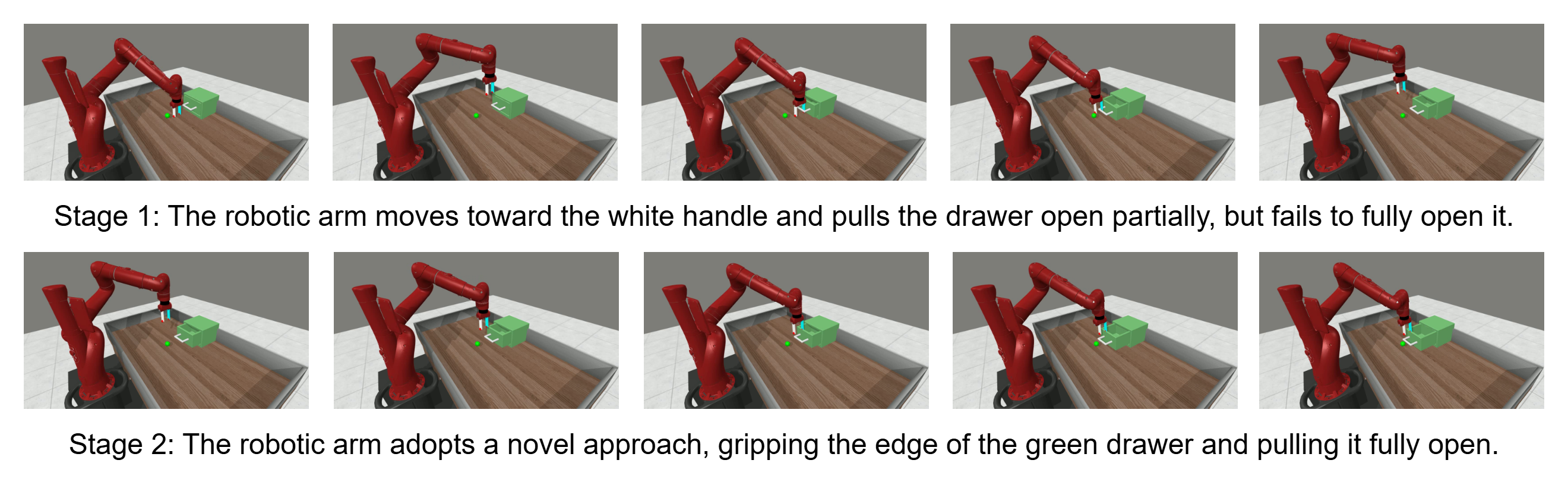}
    \vspace{-2.0em}
    \caption{Visual evidence of emergent self-correction and strategy adaptation. When an initial plan (top) fails to fully open the drawer, VICX treats the failure as informative visual context. The video generation model then functions as a reasoning engine, adaptively replanning a novel grasping strategy (bottom) to complete the task based on observed environmental feedback.}
    \label{fig:self_correction_sequences}
\end{figure}

\noindent \textbf{Emergence of Self-correction and Strategy Adaptation.} Beyond aggregate success, VICX enables the video planner to exhibit emergent self-correction. As illustrated in the \texttt{drawer-open} example (Fig. \ref{fig:self_correction_sequences}), when the initial plan fails to fully open the drawer (Stage 1), the failure is not discarded but fed back as informative visual context. In the subsequent cycle, the model does not merely repeat the failed trajectory; instead, it demonstrates an emergent strategic adaptation by targeting the drawer's edge, which is a successful grasping point that never appeared in the drawer-opening training demonstrations. This suggests that the video generation model acts as a closed-loop reasoning engine that can leverage its internalized world priors to ``invent'' novel error-recovery strategies, transitioning from semantic failure to goal completion without any hand-coded logic or task-specific fine-tuning. We provide more examples of self-correction in Appendix~\ref{app:evaluation_results}.

%% file: sections/cross_embodiment.tex
\subsection{Cross-domain Generalization}
\label{sec:result_generalization}

\noindent \textbf{Cross-task Generalization.} The ability of VICX to solve unseen tasks stems from the decoupling of task-agnostic execution and general-purpose planning:
\begin{wrapfigure}{l}{0.55\linewidth}
        \centering
        \includegraphics[width=\linewidth]{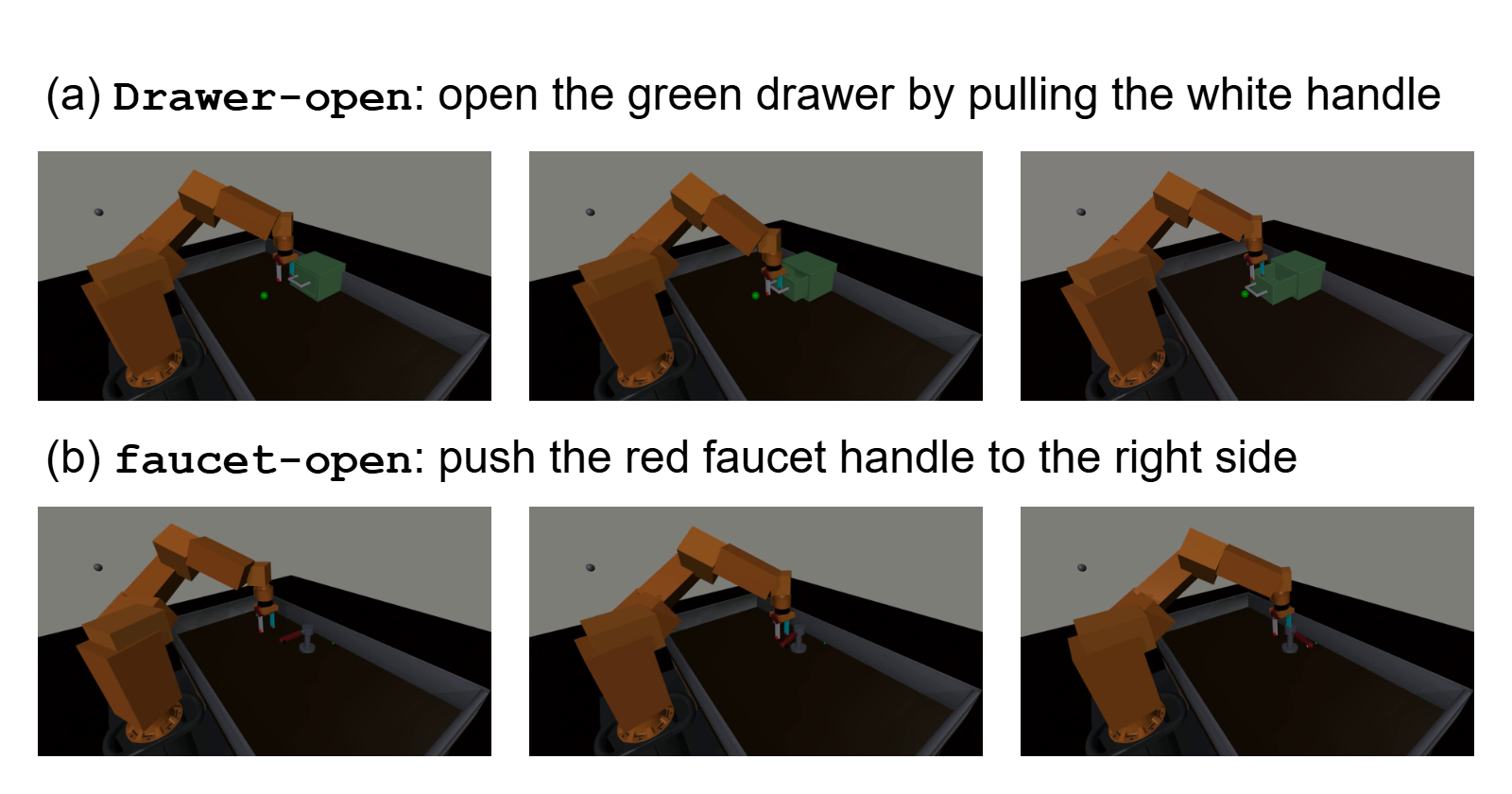}
        \caption{Cross-embodiment closed-loop evaluation of VICX. V2T-ICON in VICX is trained on the standard red Sawyer-style arm while evaluated on an unseen orange robotic arm with modified tabletop, floor texture, and lighting. (a) \texttt{drawer-open} and (b) \texttt{faucet-open} show representative task executions under this shifted condition.}
        \label{fig:cross-embodiment}
\end{wrapfigure}
\begin{itemize}[leftmargin=*]
    \item \textbf{Task-Agnostic Mapping:} V2T-ICON acts as a universal translator that focuses on robot kinematics rather than task-specific semantics. By training exclusively on arm-only images, the model is forced to ignore high-level object context (e.g., whether it is interacting with a drawer or a faucet) and instead learns the fundamental visual-to-state correspondence. This allows the V2T-ICON to generalize to unseen manipulation tasks, effectively mapping previously unseen visual plans into actionable trajectories.
    
    \item \textbf{General-Purpose World Prior:} For high-level planning, the frozen video generation model provides a rich prior for world physics \cite{wiedemer2025video,khachatryan2023text2video,wan2025wan}. Unlike models trained solely on sparse robot data, a general-purpose video model has internalized internet-scale knowledge of spatial relations and object affordances. It understands fundamental physical common sense, such as gravity, collision, and deformation, ensuring that its generated ``visual plans" are inherently physically plausible. This allows the agent to navigate the complex solution space of novel tasks in a zero-shot manner without environment-specific fine-tuning.
\end{itemize}

To further evaluate generalization, we stress-test our method on a custom DIY sweep-soccer task outside the Meta-World family, achieving zero-shot success (see Appendix \ref{app:diy_sweep_soccer}).

% \noindent \textbf{Cross-embodiment Generalization.} We next evaluate whether the same planner-translator stack remains effective when the robot embodiment and scene appearance are jointly shifted out of distribution. We keep the simulator dynamics, controller, camera viewpoint, Wan video model, and V2T checkpoint fixed, and apply the same visual variant benchmark to all nine Meta-World tasks. The shifted condition replaces the stock Sawyer shell with a boxier industrial-style arm body and simultaneously changes the Meta-World environment appearance, including the tabletop, floor texture, and scene lighting. Appendix Fig.~\ref{fig:cross_embodiment_case_studies} shows representative button-press-topdown and drawer-close case studies under this shift, illustrating how the closed-loop video agent can recover from an unsuccessful planning cycle in later replanning attempts. Because Wan conditions on both the current image and the prompt, each task uses a variant-matched prompt so that the rollout remains consistent with the shifted embodiment rather than drifting back toward the canonical scene. This should be read as a \emph{visual} generalization test rather than a dynamics-transfer test, since the underlying kinematics and gripper are unchanged. Appendix~\ref{app:variant_cross_embodiment_eval} gives the matched 5-context and 0-context evaluation protocol in detail.
\noindent \textbf{Cross-embodiment Generalization.} We further evaluate the joint shift of robot embodiment and scene appearance (Fig.~\ref{fig:cross-embodiment}). While V2T-ICON is trained on standard red robot demonstrations, it is evaluated without retraining on a boxier orange industrial arm in modified environments. As summarized in Fig.~\ref{fig:icl_context_comparison}, VICX in the 5-context setting generalizes effectively to this out-of-distribution (OOD) setting, maintaining a 57.2\% overall success rate compared to 72.2\% in the canonical setting in Sec.~\ref{sec:result:sim}. 
This performance highlights a dual-level generalization: (1) Visual-Prompt Adaptation, where the video generation model conditions its planning on the current scene to ensure visual consistency with the new embodiment; and (2) Morphological Invariance, where the V2T-ICON  aligns the motion topology of the unseen orange arm with the original red arm's demonstrations. Crucially, the model translates orange-arm queries using solely red-arm references, proving it learns structural correspondence over surface texture. This synergy enables robust OOD execution without robot-specific fine-tuning.
Detailed per-task results and protocols are provided in Appendix~\ref{app:variant_cross_embodiment_eval}.

%% file: sections/benefits_ICL.tex
\subsection{Effectiveness of In-Context Learning in V2T-ICON}
\label{sec:result:icl}

\begin{figure}[!t]
        \centering
        \includegraphics[width=\linewidth]{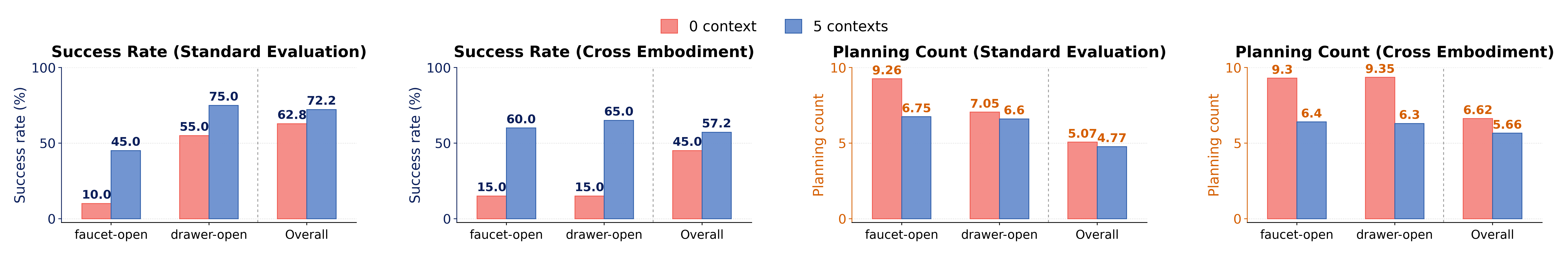}
        \caption{Overall effect of retrieved context in VICX under standard and cross-embodiment evaluation. Adding five retrieved references improves success rate (higher is better) and reduces planning count (lower is better) relative to 0-context inference in both evaluation settings.}
        \label{fig:icl_context_comparison}
\end{figure}

We evaluate the effect of the retrieved image-state context by comparing V2T-ICON with and without five retrieved references for each query frame. As summarized in Figure~\ref{fig:icl_context_comparison}, adding in-context references not only improves the overall success rate from 62.8\% to 72.2\% in the canonical environment and from 45.0\% to 57.2\% under the cross-embodiment shift, but also reduces the aggregate penalized planning count from 5.07 to 4.77 and from 6.62 to 5.66 in both settings, respectively. Detailed task-level values are reported in Tables~\ref{tab:icl_context_comparison} and \ref{tab:cross_embodiment_context}. Nevertheless, the aggregate results show that in-context references improve both closed-loop effectiveness and planning efficiency.

% We attribute these gains to the physical grounding provided by retrieved image-state references. This interpretation is consistent with prior studies of in-context learning, which show that models can use examples in the input context to construct context-dependent predictors without parameter updates~\citep{brown2020language,akyurek2023what}. In our setting, the retrieved references serve as local image-state demonstrations. Without context, V2T-ICON must infer robot state trajectories directly from generated videos, which can introduce errors when visually plausible motion does not fully align with the robot's kinematics or control space. In contrast, retrieved references provide local examples of how visual observations correspond to robot states, allowing V2T-ICON to calibrate the video-to-trajectory mapping at inference time. This in-context calibration reduces translation distortion between generated visual plans and executable robot motion. As a result, the context-based agent produces more reliable trajectories and achieves stronger closed-loop performance.

We attribute these gains to three effects of in-context calibration:
\begin{itemize}[leftmargin=*]
    \item \textbf{Inference-time physical calibration.}
    Retrieved image-state references anchor generated arm appearances to nearby robot states, allowing V2T-ICON to instantiate a query-specific visual-to-state mapping at test time instead of relying only on the mapping without parameter fine-tuning.
    \item \textbf{Cross-embodiment structural alignment.}
    Under the cross-embodiment shift, V2T-ICON translates videos of an unseen orange industrial arm using references from the original red Sawyer-style arm. The improvement from $45.0\%$ to $57.2\%$ suggests that retrieved context helps align motion structure across appearance changes, rather than relying purely on surface texture.
    \item \textbf{Contact-sensitive grounding.}
    Figure~\ref{fig:icl_context_comparison} suggests that context is most useful when small visual errors can cause large execution failures, such as \texttt{drawer-open} or \texttt{faucet-open}. In these cases, local image-state anchors reduce translation distortion between generated visual plans and executable robot trajectories.
\end{itemize}
Overall, retrieved context is not merely additional input; it provides the local physical grounding that makes V2T-ICON behave less like a global frame-wise regressor and more like an adaptive state estimator.

%% file: sections/related_work.tex
\noindent \textbf{Video Models as Explicit Planners.} Video prediction enables robots to plan explicitly in pixel space by first generating a future visual trajectory and then translating it into executable actions. Classical visual foresight used action-conditioned video models with Model Predictive Control (MPC) to reach visual goals \cite{finn2017deep,ebert2018visual}. Recent generative methods like UniPi \cite{du2023learning} and VLP \cite{du2024video} extend this to long-horizon planning by combining text-conditioned video generation with inverse dynamics or VLM-based search. Follow-up works improve execution by extracting actions from videos via dense correspondences or diffusion-based conditioning \cite{ko2024learning,wang2025language,pmlr-v270-liang25b}. Scaling this further, Large Video Planner proposed in \cite{chen2025large} enables zero-shot visual planning across novel tasks by training on massive human-robot datasets. Despite their interpretability, such planners remain limited by video fidelity, temporal consistency, physical plausibility, and the reliability of the action-extraction bridge \cite{wang2026world}.

\noindent \textbf{In-Context Learning for Robot Policy.} In-context learning bypasses parameter updates, instead adapting robot policies through conditioning on demonstrations or retrieved contexts to infer actions in novel environments. Prior works show that pretrained LLMs and VLMs can serve as general pattern learners for robot decision making \cite{mirchandani2023large} when observations and actions are represented as keypoints, action tokens, object-centric graphs, or spatial contexts \cite{dipalo2024kat,vosylius2025instant,chen2026retrieval}. Beyond manipulation, similar ideas enable policies for humanoid, quadrotor, and cross-body locomotion to adapt from proprioceptive histories or long temporal contexts \cite{radosavovic2024real,eschmann2026raptor,liu2025locoformer}. More recent robot-specific in-context policies further combine trajectory prompts or retrieval-augmented VLA models to improve few-shot adaptation \cite{fu2024incontext,sridhar2025ricl,yoo2025robossm}. These studies suggest that context provides useful task- and embodiment-specific grounding at test time, allowing policies to adapt without explicit fine-tuning.

%% file: sections/conclusion.tex
We presented VICX, a decoupled framework for generalizable robot manipulation that addresses the vision-to-execution bridge in video-based planning. 
The central components are a video generation model and a video-to-trajectory in-context operator network (V2T-ICON), a task-agnostic model that translates generated arm videos into executable robot-state trajectories rather than directly extracting low-level actions. 
By using arm-only observations and retrieved image-state prompts, V2T-ICON calibrates the visual-to-state correspondence at inference time, allowing a fixed execution module to generalize to novel task-induced trajectories and shifted robot embodiments. 
Together with closed-loop replanning, this design reduces accumulated errors and supports robust execution from generated visual plans. 
Experiments on Meta-World demonstrate cross-task generalization, self-correction, and cross-embodiment transfer, highlighting the dual generalization enabled by decoupling visual planning from task-agnostic execution grounding.

%% file: sections/limitation.tex
While demonstrating dual generalization, VICX is not yet designed for high-frequency action output. The VICX sequential pipeline, comprising video generation, trajectory translation, and low-level tracking, is better suited for short-horizon replanning than dense reactive control. Video generation remains the primary latency bottleneck; while V2T-ICON can be accelerated via quantization or caching, reducing planning latency requires future integration of distilled, streaming, or causal video models. Finally, our single-view grounding lacks the precision required for fine manipulation (e.g., millimeter-level insertion or force-sensitive contact), where minor geometric errors in either the visual plan or translated trajectory can lead to failure.

%% file: sections/appendix_exp_train_details.tex
\subsection{Tasks and Training Datasets}
\label{app:tasks_dataset}

\begin{figure}[H]
    \centering
    \includegraphics[width=\linewidth]{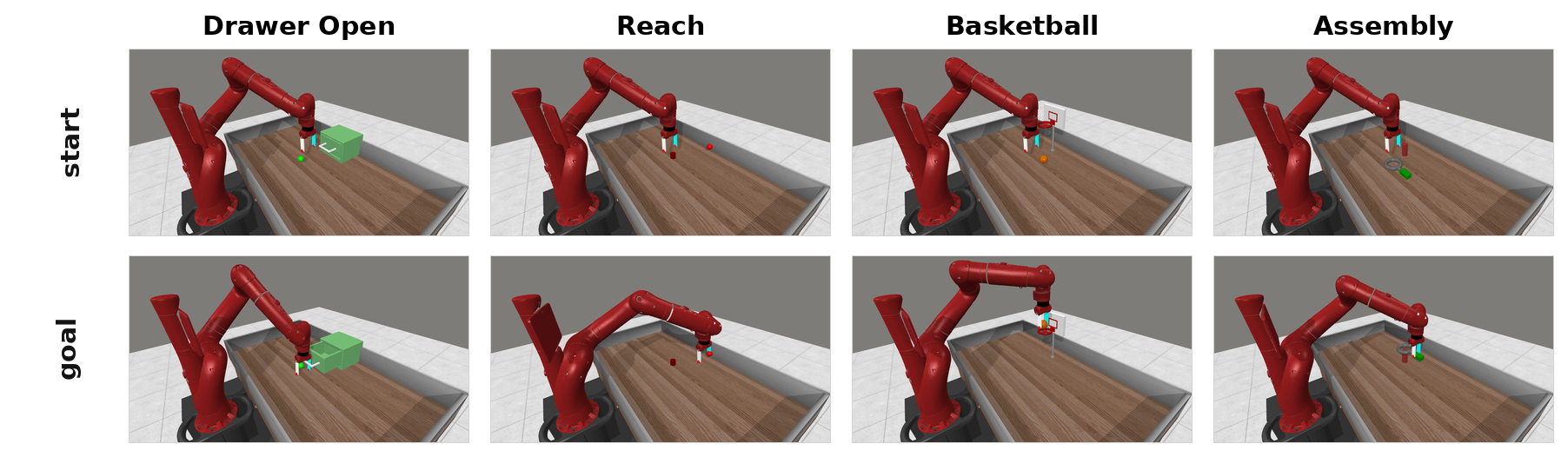}
    \caption{Overview of representative Meta-World tasks in simulation.}
    \label{fig:metaworld_tasks_overview}
\end{figure}

We conduct simulation experiments in Meta-World, a Sawyer-manipulation benchmark with 50 robotic control tasks~\citep{mclean2025metaworld}. The suite covers reaching, pushing, object transport, drawer and door manipulation, button pressing, and assembly-style insertion. This task diversity allows us to test whether a single video-conditioned execution module can translate visual plans into robot state trajectories across different objects, goals, and contact patterns. Figure~\ref{fig:metaworld_tasks_overview} shows four representative tasks from our collected trajectories.

\noindent \textbf{Observation Space.} Each collected episode stores a fixed third-person RGB stream and a compact robot state. The RGB observation is saved at $704\times1280$ resolution. In our setting, V2T-ICON uses the RGB sequence as the visual condition and predicts the corresponding state sequence. We also derive arm-only observations from the same frames. These arm-only images remove task objects and background clutter, so retrieval and trajectory translation focus on robot geometry.

\noindent \textbf{Action Space.} Meta-World uses a shared four-dimensional continuous action space across tasks. The first three dimensions command end-effector displacement in Cartesian space. The fourth dimension controls the gripper. The V2T-ICON translator predicts state trajectories rather than actions. A low-level controller then tracks the predicted states during closed-loop evaluation.

\noindent \textbf{Data Collection.} We collect demonstrations by rolling out the scripted expert policies provided by Meta-World from randomized initial states. For each rollout, we render the fixed camera view and record the four-dimensional robot state at every timestep. After collecting the raw trajectories, we apply SAM3-based whole-arm segmentation to create the arm-only frames used by V2T-ICON.

\begin{table}[H]
    \centering
    \small
    \caption{Source demonstrations used for V2T-ICON training.}
    \label{tab:source_dataset_stats}
    \begin{tabular}{lrr}
        \hline
        Task & Episodes & Frames \\
        \hline
        \texttt{reach} & 950 & 42,859 \\
        \texttt{drawer-open} & 200 & 17,717 \\
        \texttt{basketball} & 200 & 18,470 \\
        \hline
        Total & 1,350 & 79,046 \\
        \hline
    \end{tabular}
\end{table}

For the main experiments, V2T-ICON is trained on three source tasks: \texttt{reach}, \texttt{drawer-open}, and \texttt{basketball}. These tasks provide 1,350 training episodes and 79,046 action steps, as summarized in Table~\ref{tab:source_dataset_stats}. They cover direct reaching, contact-rich pulling, and grasp-lift-release behavior. At evaluation time, the trained translator is reused without task-specific fine-tuning on the nine closed-loop tasks reported in Sec.~\ref{sec:result:sim}.

\subsection{V2T-ICON Training}
\label{app:v2t_icon_training}

\noindent \textbf{Model Architecture.} V2T-ICON learns a retrieval-conditioned video-to-trajectory operator. Given a query video window $Q=\{I_t^q\}_{t=1}^{T}$ and retrieved context $R=\{R_t\}_{t=1}^{T}$, where $R_t=\{(I_{t,j}^r,x_{t,j}^r)\}_{j=1}^{N}$ contains image-state references for frame $t$, the model predicts an executable state trajectory
\[
\hat{X}=\mathcal{G}_{\theta}(Q,R)=\{\hat{x}_t\}_{t=1}^{T}.
\]
Thus, the inputs are the query frames and their retrieved visual-state context, and the output is one executable robot state for each query frame. V2T-ICON consists of the following components:

\begin{itemize}
    \item \textbf{Image encoder:} A frozen DINOv2 visual encoder maps each query and reference image into patch-level features and a global image feature~\citep{oquab2024dinov}.
    \item \textbf{State encoder:} A learned linear encoder maps each retrieved robot state into the shared hidden dimension used by the in-context attention blocks.
    \item \textbf{RoPE embedding:} Head-wise rotary position embeddings encode set and temporal positions, allowing attention to reason over retrieved examples and ordered video frames~\citep{su2024roformer}.
    \item \textbf{Reference attention:} A per-frame in-context attention module compares the query image with its retrieved image-state references and produces a frame latent.
    \item \textbf{Temporal attention:} A non-causal temporal attention module propagates information across all frame latents to enforce trajectory-level consistency.
    \item \textbf{Trajectory decoder:} A linear decoder maps the temporally contextualized hidden states to the predicted robot-state trajectory.
\end{itemize}

\noindent \textbf{Input Encoding.} All query and reference images are encoded by a frozen visual encoder $\phi_{\mathrm{vis}}$. For each query image and reference image, we compute
\[
(P_t^q,c_t^q)=\phi_{\mathrm{vis}}(I_t^q),\qquad
(P_{t,j}^r,c_{t,j}^r)=\phi_{\mathrm{vis}}(I_{t,j}^r),
\]
where $P$ denotes patch-level features and $c$ denotes a global image feature. A learned state encoder $\psi_x$ maps each retrieved state into the shared hidden dimension. We write the query token and reference tokens as
\[
e_t^q=\eta_q(P_t^q,c_t^q),\qquad
e_{t,j}^r=\eta_r(P_{t,j}^r,c_{t,j}^r,\psi_x(x_{t,j}^r)),
\]
where $\eta_q$ and $\eta_r$ are learned projection layers that align visual and state features into the Transformer hidden space.

\noindent \textbf{Reference Attention.} The first stage operates independently at each time step. At refinement round $k$, reference attention compares a query-side token $q_t^{(k)}$ with the retrieved reference tokens for the same frame and produces a context-grounded frame latent:
\[
h_t^{(k)}
=
f_{\mathrm{ref}}\left(q_t^{(k)},\{e_{t,j}^r\}_{j=1}^{N}\right),
\qquad
h_t^{(k)}\in\mathbb{R}^{d}.
\]
For the first round, $q_t^{(1)}=e_t^q$. In later rounds, $q_t^{(k)}$ is formed by fusing the original query image token with the previous temporal hidden state. Because every reference token includes both image evidence and a paired robot state, this module estimates the query state through visual correspondence to local examples rather than through direct regression from the query image alone.

\noindent \textbf{Temporal Attention.} The per-frame latents are stacked into a video-level sequence,
\[
H^{(k)}=[h_1^{(k)},\ldots,h_T^{(k)}],
\]
and processed by a non-causal temporal attention module:
\[
Z^{(k)}
=
f_{\mathrm{temp}}(H^{(k)}),
\qquad
Z^{(k)}=\{z_t^{(k)}\}_{t=1}^{T}.
\]
Temporal attention lets each frame use information from the entire query window, helping resolve ambiguous single-frame arm poses, reduce high-frequency prediction noise, and exploit the smoothness of physical robot motion.

\noindent \textbf{Iterative Refinement.} We repeat the reference-attention and temporal-attention modules for $K$ refinement rounds. The first round starts from the encoded query image tokens:
\[
H^{(1)}
=
\left\{
f_{\mathrm{ref}}\left(e_t^q,\{e_{t,j}^r\}_{j=1}^{N}\right)
\right\}_{t=1}^{T},
\qquad
Z^{(1)}=f_{\mathrm{temp}}(H^{(1)})\triangleq\{z_t^{(1)}\}_{t=1}^{T}.
\]
For later rounds, the previous temporal representation is used as a query-side prior. Here $f_{\mathrm{fuse}}$ is a learned fusion layer that combines the original query token and the previous temporal hidden state, implemented as a projection from their concatenation back to the Transformer hidden dimension:
\[
q_t^{(k)}=f_{\mathrm{fuse}}(e_t^q,z_t^{(k-1)})
=W_{\mathrm{fuse}}[e_t^q;z_t^{(k-1)}]+b_{\mathrm{fuse}},
\]
\[
H^{(k)}
=
\left\{
f_{\mathrm{ref}}\left(q_t^{(k)},\{e_{t,j}^r\}_{j=1}^{N}\right)
\right\}_{t=1}^{T},
\]
\[
Z^{(k)}=f_{\mathrm{temp}}(H^{(k)})\triangleq\{z_t^{(k)}\}_{t=1}^{T},
\quad k=2,\ldots,K.
\]
The trajectory decoder is applied only after the final refinement round:
\[
\hat{X}=f_{\mathrm{dec}}(Z^{(K)}).
\]
This design lets V2T-ICON first form a rough state estimate from retrieved examples and then refine it using trajectory-level context before producing the final executable state sequence.

Algorithm~\ref{alg:v2t_icon_forward} summarizes the model forward computation.

\begin{algorithm}[H]
    \caption{V2T-ICON model forward pass.}
    \label{alg:v2t_icon_forward}
    \small
    \begin{algorithmic}[1]
        \Require Query video window $Q=\{I_t^q\}_{t=1}^{T}$, retrieved context $R=\{R_t\}_{t=1}^{T}$ with $R_t=\{(I_{t,j}^r,x_{t,j}^r)\}_{j=1}^{N}$, refinement rounds $K$
        \Ensure Predicted robot-state trajectory $\hat{X}=\{\hat{x}_t\}_{t=1}^{T}$
        \For{$t=1,\ldots,T$}
            \State $(P_t^q,c_t^q)\gets\phi_{\mathrm{vis}}(I_t^q)$
            \State $e_t^q\gets\eta_q(P_t^q,c_t^q)$
            \For{$j=1,\ldots,N$}
                \State $(P_{t,j}^r,c_{t,j}^r)\gets\phi_{\mathrm{vis}}(I_{t,j}^r)$
                \State $s_{t,j}^r\gets\psi_x(x_{t,j}^r)$
                \State $e_{t,j}^r\gets\eta_r(P_{t,j}^r,c_{t,j}^r,s_{t,j}^r)$
            \EndFor
            \State $q_t^{(1)}\gets e_t^q$
        \EndFor
        \For{$k=1,\ldots,K$}
            \For{$t=1,\ldots,T$}
                \State $h_t^{(k)}\gets f_{\mathrm{ref}}\left(q_t^{(k)},\{e_{t,j}^r\}_{j=1}^{N}\right)$
            \EndFor
            \State $Z^{(k)}\gets f_{\mathrm{temp}}\left([h_1^{(k)},\ldots,h_T^{(k)}]\right)$ \Comment{$Z^{(k)}=\{z_t^{(k)}\}_{t=1}^{T}$}
            \If{$k<K$}
                \For{$t=1,\ldots,T$}
                    \State $q_t^{(k+1)}\gets f_{\mathrm{fuse}}(e_t^q,z_t^{(k)})$
                \EndFor
            \EndIf
        \EndFor
        \State $\hat{X}\gets f_{\mathrm{dec}}(Z^{(K)})$
        \State \Return $\hat{X}$
    \end{algorithmic}
\end{algorithm}

The model is trained with a weighted trajectory objective,
$$
\mathcal{L}_{\mathrm{V2T}}
=
\lambda_{\mathrm{state}}\mathcal{L}_{\mathrm{state}}
+
\lambda_{\mathrm{smooth}}\mathcal{L}_{\mathrm{smooth}}
+
\lambda_{\mathrm{vel}}\mathcal{L}_{\mathrm{vel}} .
$$
The state term is a point-wise MSE loss,
$$
\mathcal{L}_{\mathrm{state}}
=
\frac{1}{T}\sum_{t=1}^{T}\|\hat{x}_t-x_t\|_2^2 .
$$
The smoothness term penalizes second differences of the predicted xyz trajectory,
$$
\mathcal{L}_{\mathrm{smooth}}
=
\frac{1}{T-2}\sum_{t=2}^{T-1}
\left\|
\hat{p}_{t+1}-2\hat{p}_{t}+\hat{p}_{t-1}
\right\|_2^2 .
$$
The velocity-consistency term matches predicted and target xyz velocities,
$$
\mathcal{L}_{\mathrm{vel}}
=
\frac{1}{T-1}\sum_{t=1}^{T-1}
\left\|
(\hat{p}_{t+1}-\hat{p}_t)-(p_{t+1}-p_t)
\right\|_2^2 .
$$
Here $p_t$ and $\hat{p}_t$ denote the xyz components of $x_t$ and $\hat{x}_t$, respectively.

\noindent \textbf{Training Protocol and Hyperparameters.} Training samples 25-frame windows from the arm-only demonstration pool described in Table~\ref{tab:source_dataset_stats}. For each query frame, V2T-ICON receives five retrieved image-state references from the same arm-only pool. We normalize states with percentile-based bounds before optimization. The reported model was trained on two NVIDIA H200 GPUs and required approximately 3.5 hours of wall-clock time. The remaining architecture and optimization settings are summarized in Table~\ref{tab:v2t_icon_training_hparams}.

\begin{table}[H]
    \centering
    \small
    \caption{V2T-ICON architecture and training hyperparameters.}
    \label{tab:v2t_icon_training_hparams}
    \begin{tabular}{lp{0.62\linewidth}}
        \hline
        Parameter & Value \\
        \hline
        Visual encoder & Frozen DINOv2 \\
        Query window length & 25 frames \\
        References per query frame & 5 retrieved image-state pairs \\
        Hidden dimension & 256 \\
        Reference-attention depth & 4 layers, 8 heads \\
        Temporal-attention depth & 4 layers, 8 heads, non-causal attention \\
        Refinement rounds & 3 rounds \\
        Dropout & 0.0 \\
        Optimizer & Muon, learning rate $1\times10^{-4}$, weight decay 0.01 \\
        Scheduler & 10\% warmup followed by cosine decay to 0.1 of the initial learning rate \\
        Batch size & 8 per device \\
        Training length & 8,000 maximum steps \\
        Gradient clipping & 1.0 \\
        Loss weights & state 1.0, smoothness 0.05, velocity consistency 0.05 \\
        \hline
    \end{tabular}
\end{table}

%% file: sections/appendix_exp_eval_details.tex
\subsection{VICX Manipulation Pseudocode}
\label{app:video_agent_pseudocode}

Algorithm~\ref{alg:video_agent_eval} gives the evaluation procedure corresponding to Fig.~\ref{fig: closed_loop_evaluation}. The VICX interface supports both expert-video open-loop execution and video-agent closed-loop replanning. In the open-loop setting, the planned video can be supplied externally, for example from an expert demonstration. In the closed-loop setting, the video agent is queried from the current observation at every planning cycle. We use $\mathcal{V}$ for the frozen video agent, $\mathcal{G}_{\theta}$ for V2T-ICON, $\mathcal{S}$ for the arm-only segmentation operator, $Q=\{I_t^q\}_{t=1}^{T}$ for a planned video, and $R=\{R_t\}_{t=1}^{T}$ for the retrieved reference context. For each planned video, we retrieve $R$ from the arm-only training pool $\mathcal{D}_{\text{train}}$ as in Sec.~\ref{sec: method}, and V2T-ICON predicts the executable state trajectory $\hat{X}=\mathcal{G}_{\theta}(Q,R)$.

\begin{algorithm}[H]
    \caption{VICX via video generation and V2T-ICON.}
    \label{alg:video_agent_eval}
    \small
    \begin{algorithmic}[1]
        \Require Task prompt $P$, initial observation $I_0$, planning budget $C$, reference count $N$, arm-only training pool $\mathcal{D}_{\text{train}}$, optional external planned video $Q_{ext}$
        \Ensure Planned videos, executable state trajectories, and success flag $y$
        \State $I \gets I_0$
        \If{$C = 0$} \Comment{open-loop execution}
            \State $Q \gets Q_{ext}$ if $Q_{ext}$ is provided; otherwise $Q \gets \mathcal{V}(P,I)$
            \State $R \gets \operatorname{RetrieveRefs}(\mathcal{S}(Q), \mathcal{D}_{\text{train}}, N)$
            \State $\hat{X} \gets \mathcal{G}_{\theta}(Q,R)$
            \State $y \gets \operatorname{Execute}(\hat{X})$
            \State \Return $Q, \hat{X}, y$
        \EndIf
        \For{$c = 1,\ldots,C$} \Comment{closed-loop replanning}
            \State $Q_c \gets \mathcal{V}(P,I)$
            \State $R_c \gets \operatorname{RetrieveRefs}(\mathcal{S}(Q_c), \mathcal{D}_{\text{train}}, N)$
            \State $\hat{X}_c \gets \mathcal{G}_{\theta}(Q_c,R_c)$
            \State $y_c, I^c_{t:t+T} \gets \operatorname{ExecuteAndObserve}(\hat{X}_c)$
            \State $I \leftarrow I^c_{t:t+T}$
            \If{$y_c = \operatorname{success}$}
                \State \Return $\{Q_i,\hat{X}_i\}_{i=1}^{c}, \operatorname{success}$
            \EndIf
        \EndFor
        \State \Return $\{Q_i,\hat{X}_i\}_{i=1}^{C}, \operatorname{failed}$
    \end{algorithmic}
\end{algorithm}

\subsection{Evaluation Metrics}
\label{app:evaluation_metrics}

We report two metrics for VICX manipulation: success rate and penalized planning count. Consider a fixed task and method setting evaluated over $M$ independent runs. Each run has a planning budget of $C$ cycles, with $C=10$ in our experiments. For run $m$, let $y_m \in \{0,1\}$ denote the final task-completion indicator, where $y_m=1$ if the task succeeds within the budget and $y_m=0$ otherwise. For successful runs, let $c_m \in \{1,\ldots,C\}$ denote the first planning cycle at which success is achieved.

\noindent \textbf{Success Rate.} Success rate measures the empirical probability of completing the task within the planning budget:
$$
\operatorname{SR}
= \frac{1}{M}\sum_{m=1}^{M} y_m .
$$
Higher values indicate better task performance.

\noindent \textbf{Penalized Planning Count.} Penalized planning count measures how many planning cycles are required while assigning failed runs the full budget. For each run, we define the penalized cycle count as
$$
\tilde{c}_m =
\begin{cases}
c_m, & \text{if } y_m = 1,\\
C, & \text{if } y_m = 0.
\end{cases}
$$
The penalized planning count is then
$$
\operatorname{PPC}
= \frac{1}{M}\sum_{m=1}^{M} \tilde{c}_m .
$$
Lower values indicate more efficient closed-loop inference. Unlike success rate alone, this metric captures both early termination in successful runs and the cost of unsuccessful runs that exhaust the replanning budget. For aggregate results across a set of tasks, we compute both metrics over the union of all evaluated runs.

\subsection{Evaluation Settings and Results}
\label{app:evaluation_results}

\noindent \textbf{Baseline.} We evaluate three external baselines using the default checkpoints and launch settings provided by the evaluation scripts.
\begin{itemize}
    \item \textbf{$\pi_{0.5}$-Scratch.} $\pi_{0.5}$ is a Vision-Language-Action model from Physical Intelligence designed for open-world generalization and broadly used as a generalist policy for robotic manipulation tasks. Since the official $\pi_{0.5}$ model is closed-source and not available as a directly reproducible public checkpoint, we use the open checkpoint from the LeRobot implementation\footnote{\url{https://huggingface.co/docs/lerobot/pi05}} as the baseline evaluation model.
    \item \textbf{$\pi_{0.5}$-Finetune.} To reduce the effect of evaluating a scratch $\pi_{0.5}$ policy on Meta-World scenes that it has not observed, we also evaluate a fine-tuned checkpoint. This checkpoint is initialized from $\pi_{0.5,\text{base}}$ and then finetuned on the LeRobot Meta-World MT50 dataset\footnote{\url{https://huggingface.co/datasets/lerobot/metaworld_mt50}}. The Hugging Face dataset card for this dataset reports 2500 episodes, 204806 frames, and 49 tasks.
    \item \textbf{AVDC.} AVDC is a hierarchical manipulation framework that combines future video foresight with action generation from the predicted visual dynamics. This design is closely related to VICX, where future visual plans are translated into executable robot trajectories. We therefore include AVDC as a comparison method. We use the official Meta-World video diffusion checkpoint \texttt{ckpts/metaworld/model-24.pt}\footnote{\url{https://huggingface.co/Po-Chen/flowdiffusion}} released by Ko et al.~\cite{ko2024learning}. Following the AVDC Meta-World setup, this checkpoint is trained on 11 Meta-World tasks, with videos rendered from three camera poses and five demonstrations per task per camera pose, for a total of 165 videos.
\end{itemize}

We note that a strictly controlled apples-to-apples comparison between these systems is difficult because the methods instantiate different interfaces between perception, planning, and control. End-to-end VLA baselines directly predict robot actions, AVDC combines video foresight with action generation, whereas our system evaluates a closed-loop video-to-trajectory pipeline in which a frozen video model proposes visual plans and V2T-ICON grounds them into state trajectories using retrieved references. Equalizing one factor would therefore change another: removing closed-loop replanning or in-context grounding from our system would no longer evaluate the proposed interface, while adding the same planner or retrieval mechanism to the baselines would create new hybrid methods rather than their original algorithms. In addition, available public checkpoints differ in training data, implementation, and release status. We therefore view Table~\ref{tab:sim_main_results} as a reproducible reference comparison under available public implementations.

\noindent \textbf{5 Contexts Evaluation Setting.} The evaluation reported in Sec.~\ref{sec:result:sim} and  Table~\ref{tab:sim_main_results} instantiates Algorithm~\ref{alg:video_agent_eval} in the closed-loop setting. We use Wan 2.7 as the frozen video agent $\mathcal{V}$, a trained V2T-ICON checkpoint as $\mathcal{G}_{\theta}$, a planning budget of $C=10$, and $N=5$ retrieved image-state references per query frame. At each planning cycle, Wan is prompted from the current Meta-World observation, the generated video is segmented into arm-only query frames, V2T-ICON predicts an executable state trajectory from the planned video and retrieved references, and the predicted trajectory is executed in the simulator. The run stops early when the task succeeds; otherwise, it is counted as a failure after all 10 planning cycles are exhausted.

The task prompt $P$ in Algorithm~\ref{alg:video_agent_eval} is task specific and is passed directly to Wan. Figure~\ref{fig:wan_drawer_open_prompt} shows the prompt used for \texttt{drawer-open} as a representative example.

\begin{figure}[!t]
    \centering
    \definecolor{wanpromptheader}{gray}{0.50}
    \definecolor{wanpromptbody}{gray}{0.94}
    \begingroup
    \setlength{\fboxsep}{0pt}
    \setlength{\fboxrule}{0.4pt}
    \fbox{%
        \begin{minipage}{0.92\linewidth}
            \noindent\colorbox{wanpromptheader}{%
                \parbox{\linewidth}{%
                    \vspace{0.6ex}
                    \hspace{0.9em}\textcolor{white}{\textbf{Wan Prompt Example: \texttt{drawer-open}}}
                    \vspace{0.6ex}
                }%
            }\par\nointerlineskip
            \noindent\colorbox{wanpromptbody}{%
                \parbox{\linewidth}{%
                    \vspace{1.0ex}
                    \hspace{1em}%
                    \begin{minipage}{0.94\linewidth}
                        \textbf{Goal:} Open the green drawer by pulling the white handle straight outward until the pulled-out drawer front/handle reaches and touches the fixed green ball on the tabletop.

                        \vspace{0.8ex}
                        \textbf{Scene:} Fixed corner-camera view of a red Sawyer arm with a parallel-jaw gripper above a wooden tabletop. A green drawer cabinet sits on the right side of the table with a white horizontal handle protruding from the drawer front. A small green ball may be visible on the table but is not relevant. Continue from the current visual state while preserving the existing camera geometry, object identities, and any progress already completed.

                        \vspace{0.8ex}
                        \textbf{Priority order:}
                        \begin{enumerate}
                            \setlength{\itemsep}{0.25ex}
                            \setlength{\parsep}{0pt}
                            \setlength{\topsep}{0.4ex}
                            \item Begin motion in the first frame. Keep the gripper's initial finger spacing fixed, with no opening, closing, pulsing, or aperture change, and move immediately to a position directly above the middle of the white handle using a short, direct approach, without circling or hovering.
                            \item Descend vertically all the way to the lowest contact position before any outward pull begins. The white handle must be clearly centered in the empty gap between the two fingers, and the end effector must complete the full downward insertion before the drawer starts moving.
                            \item Establish a solid, non-penetrating hooked contact from behind the handle while preserving the same fixed finger gap. The fingers may straddle the handle, but they must never open, close, overlap, merge with, or pass through the handle geometry. The wrist, palm, and gripper body must stay outside the drawer front and must not enter the drawer cavity.
                            \item Only after that completed downward contact is achieved, pull the drawer straight outward in a pure translational motion while preserving that clean contact geometry. The drawer must slide horizontally along its rails without any tilting, dropping, rotating, or sideways slip, then stop once it is substantially open and stable.
                        \end{enumerate}
                    \end{minipage}
                    \vspace{1.0ex}
                }%
            }%
        \end{minipage}%
    }
    \endgroup
    \caption{Full Wan task prompt used for the \texttt{drawer-open} closed-loop evaluation.}
    \label{fig:wan_drawer_open_prompt}
\end{figure}

\noindent \textbf{0 Contexts Evaluation Setting.} The no-context setting in Sec.~\ref{sec:result:icl} and Table~\ref{tab:icl_context_comparison} uses the same closed-loop evaluation protocol as Table~\ref{tab:sim_main_results}, except that the number of retrieved references is set to $N=0$ instead of $N=5$. Table~\ref{tab:icl_context_comparison} reports the task-level comparison between V2T-ICON with five retrieved references and this no-context variant.

\begin{figure}[h]
    \centering
    \includegraphics[width=\linewidth]{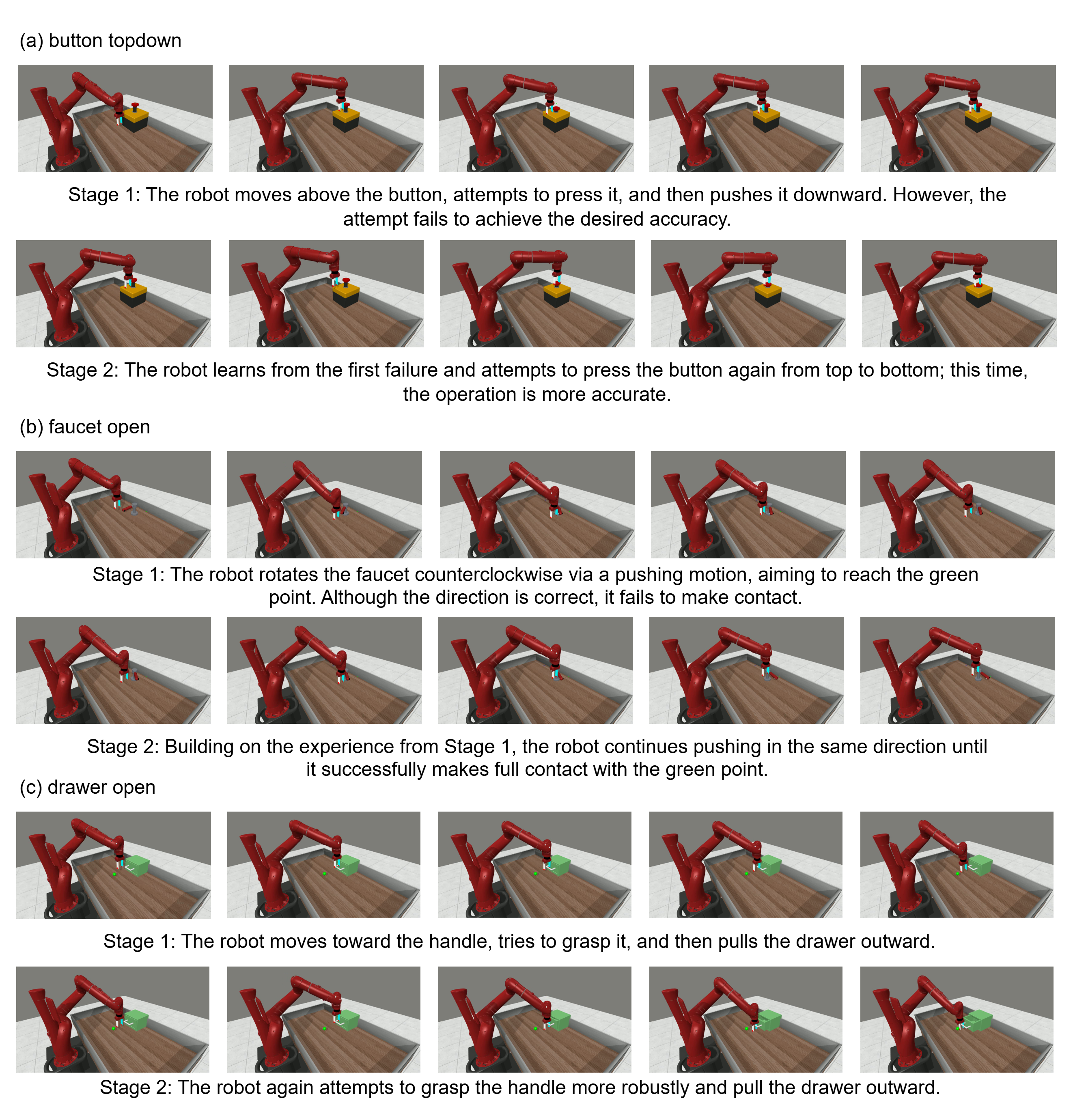}
    \caption{Additional examples of emergent self-correction and adaptive strategy results in VICX manipulation on \texttt{button-press-topdown} and \texttt{faucet-open}. }
    \label{fig:add_self_correction_sequences}
\end{figure}

\begin{figure}[h]
    \centering
    \includegraphics[width=\linewidth]{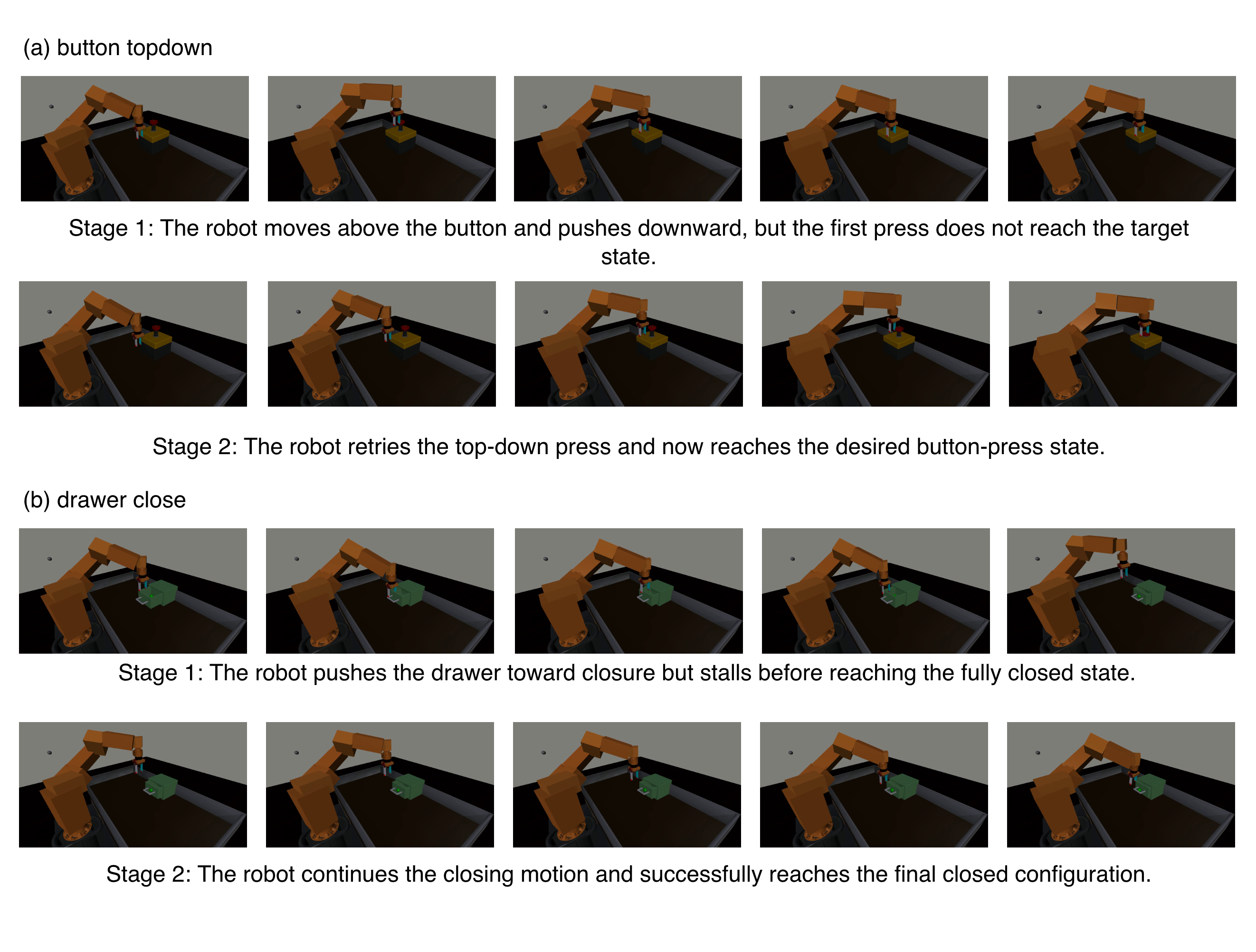}
    \caption{Representative cross-embodiment case studies under the shifted arm-body and Meta-World environment condition. Each row shows five frames sampled within a single planning cycle. The failure examples are taken from planning cycles that do not yet reach the target state, while the paired success examples show the corresponding successful planning cycles that recover under the same shifted benchmark. For \texttt{button-press-topdown}, the success case reaches the intended press state after an earlier failed top-down attempt. For \texttt{drawer-close}, the failure case stalls before full closure whereas the success case reaches the final closed state. Together, these paired cycle-level examples illustrate the self-correction behavior enabled by VICX behind the aggregate benchmark reported in Sec.~\ref{sec:result_generalization}.}
    \label{fig:cross_embodiment_case_studies}
\end{figure}

\noindent \textbf{Additional Self-correction Results.} Here, we provide further qualitative evidence to support our observations in Sec.~\ref{sec:result:sim} regarding the emergent self-correction capabilities of VICX. As illustrated in Figure~\ref{fig:add_self_correction_sequences}, when the robot encounters suboptimal execution, such as imprecise contact or gripper slippage, it does not persistently fail. Instead, the frozen video model in VICX observes the resulting state and generates a ``corrective" visual plan in the subsequent cycle. For example, in \texttt{button-press-topdown}, the agent compensates for an initial offset by re-centering its approach, while in \texttt{faucet-open}, it adjusts its grasping pose to maintain stable contact after an initial slip. These results reinforce that VICX can effectively leverage the generative priors of the video model to recover from ``out-of-distribution" failure states without any explicit self-correction supervision during training. A similar corrective pattern also appears under the cross-embodiment shift in Figure~\ref{fig:cross_embodiment_case_studies}, where later planning cycles recover from initially unsuccessful cycles despite the altered arm-body and scene appearance.

\begin{table}[!t]
    \centering
    \small
    \caption{Closed-loop inference comparison with and without context. The five-context setting uses retrieved image-state references for each query frame, while the no-context setting removes these references from the same closed-loop inference stack. \texttt{drawer-open} is included in the training dataset of V2T-ICON; all other tasks are out-of-distribution tasks.}
    \label{tab:icl_context_comparison}
    \begin{tabular}{lcccc}
        \hline
        & \multicolumn{2}{c}{Success Rate $\uparrow$} & \multicolumn{2}{c}{Penalized Planning Count $\downarrow$} \\
        \cline{2-5}
        Task & 5 Contexts & No Context & 5 Contexts & No Context \\
        \hline
        \texttt{button-press} & \textbf{20.0\%} & 0.0\% & \textbf{8.35} & 10.00 \\
        \texttt{button-press-topdown} & \textbf{85.0\%} & 70.0\% & \textbf{4.32} & 4.60 \\
        \texttt{coffee-button} & 100.0\% & 100.0\% & 1.80 & \textbf{1.20} \\
        \texttt{door-close} & 100.0\% & 100.0\% & 1.00 & 1.00 \\
        \texttt{drawer-close} & 100.0\% & 100.0\% & 3.40 & \textbf{3.15} \\{\texttt{drawer-open}} & \textbf{75.0\%} & 55.0\% & \textbf{6.60} & 7.05 \\
        \texttt{faucet-close} & 25.0\% & \textbf{30.0\%} & 9.65 & \textbf{8.50} \\
        \texttt{faucet-open} & \textbf{45.0\%} & 10.0\% & \textbf{6.75} & 9.26 \\
        \texttt{handle-press} & 100.0\% & 100.0\% & \textbf{1.00} & 1.10 \\
        \hline
        Overall & \textbf{72.2\%} & 62.8\% & \textbf{4.77} & 5.07 \\
        \hline
    \end{tabular}
\end{table}

\subsection{Cross-Embodiment Variant Evaluation}
\label{app:variant_cross_embodiment_eval}

The cross-embodiment benchmark in Sec.~\ref{sec:result_generalization} evaluates the same closed-loop stack under a matched visual variant that changes both robot appearance and scene rendering while keeping the underlying task dynamics fixed. Specifically, we replace the stock Sawyer shell with a boxier industrial-orange arm and simultaneously change the Meta-World environment appearance, including the tabletop, floor texture, and lighting. The evaluation still uses the same settings with Appendix~\ref{app:evaluation_results}, including the \texttt{corner4} camera, Wan 2.7 as the frozen video generation model, the same V2T-ICON checkpoint as the main closed-loop benchmark, and a planning budget of $C=10$ cycles.

We run this benchmark on nine Meta-World tasks: \texttt{door-close}, \texttt{drawer-close}, \texttt{drawer-open}, \texttt{button-press-topdown}, \texttt{button-press}, \texttt{faucet-close}, \texttt{faucet-open}, \texttt{handle-press}, and \texttt{coffee-button}. Each task is evaluated over 20 seeds, giving 180 runs per benchmark. The reported success rate and penalized planning count follow Appendix~\ref{app:evaluation_metrics}, with failed runs assigned the full 10-cycle budget.

\begin{figure}[h]
    \centering
    \includegraphics[width=\linewidth]{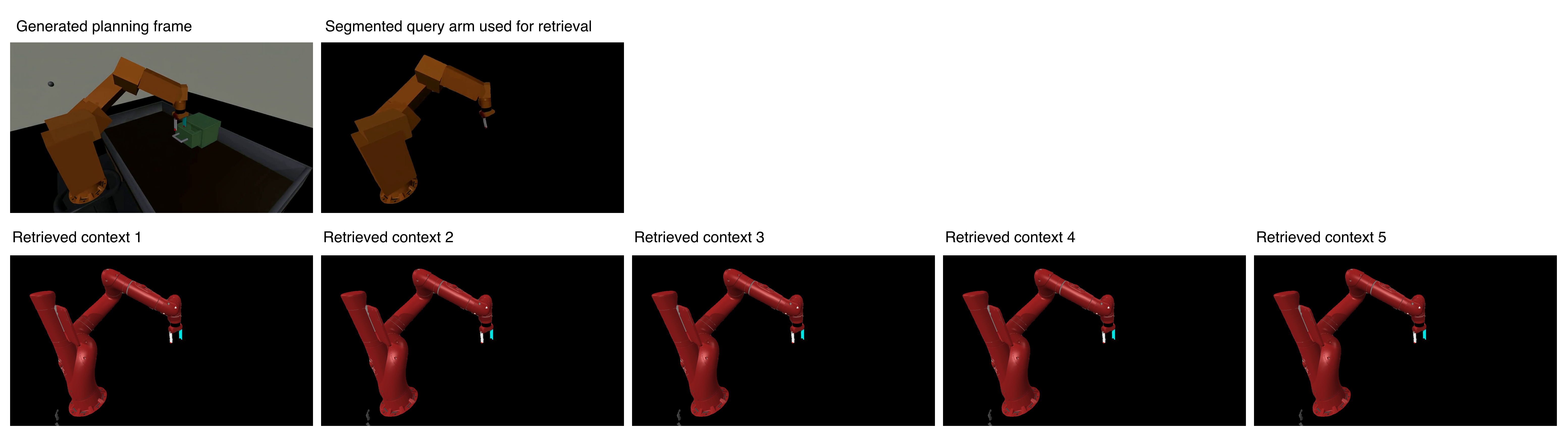}
    \caption{Retrieval illustration under the cross-embodiment benchmark. The top panels show one generated planning frame and its segmented whole-arm query image under the shifted arm-body and scene condition. The bottom panels show the top retrieved whole-arm references used by V2T-ICON at inference time. This example illustrates that retrieval is driven by arm-pose alignment rather than task identity or scene appearance, which is consistent with the task-agnostic design of the retrieval space.}
    \label{fig:cross_embodiment_retrieval_demo}
\end{figure}

\noindent \textbf{5 Contexts Evaluation Setting.} The main cross-embodiment result follows the same 5-context closed-loop procedure described in Appendix~\ref{app:evaluation_results}, now applied under the shifted embodiment-and-scene condition with $N=5$ retrieved image-state references per query frame. Figure~\ref{fig:cross_embodiment_retrieval_demo} shows a representative retrieval process from this setting, including the generated planning frame, the segmented whole-arm query image, and the top retrieved references used by V2T-ICON at inference time. This qualitative example helps interpret the 5-context benchmark by showing that retrieval remains organized by arm-pose alignment under the shifted embodiment rather than by task identity or scene appearance.

\noindent \textbf{0 Contexts Evaluation Setting.} The zero-context setting in Sec.~\ref{sec:result_generalization} and Table~\ref{tab:cross_embodiment_context} uses the same cross-embodiment evaluation protocol, task set, seeds, planning budget, video model, and V2T-ICON checkpoint as the 5-context benchmark, but removes retrieval during inference by setting the number of references to $N=0$. This gives a matched comparison between retrieval-conditioned inference and no-context inference under the same arm-body and environment shift.

\begin{table}[!t]
    \centering
    \small
    \caption{Cross-embodiment closed-loop comparison with and without inference context. Both settings use the same nine-task benchmark, ten-cycle budget, and the same arm-body and environment shift. The 5-context setting retrieves five image-state references per query frame, whereas the 0-context setting disables retrieval during inference while keeping the same V2T-ICON checkpoint fixed. PPC denotes the penalized planning count from Appendix~B.2; under this evaluation protocol it is equal to the mean completed planning cycles because failed runs consume the full 10-cycle budget.}
    \begin{tabular}{lcccc}
        \hline
        & \multicolumn{2}{c}{Success Rate $\uparrow$} & \multicolumn{2}{c}{Penalized Planning Count $\downarrow$} \\
        \cline{2-5}
        Task & 5 Contexts & No Context & 5 Contexts & No Context \\
        \hline
        \texttt{door-close} & \textbf{100.0\%} & \textbf{100.0\%} & \textbf{1.00} & 1.65 \\
        \texttt{drawer-close} & 80.0\% & \textbf{85.0\%} & 4.60 & \textbf{3.60} \\{\texttt{drawer-open}} & \textbf{65.0\%} & 15.0\% & \textbf{6.30} & 9.35 \\
        \texttt{button-press-topdown} & 25.0\% & \textbf{35.0\%} & \textbf{7.90} & 8.00 \\
        \texttt{button-press} & \textbf{0.0\%} & \textbf{0.0\%} & \textbf{10.00} & \textbf{10.00} \\
        \texttt{faucet-close} & 0.0\% & \textbf{5.0\%} & 10.00 & \textbf{9.75} \\
        \texttt{faucet-open} & \textbf{60.0\%} & 15.0\% & \textbf{6.40} & 9.30 \\
        \texttt{handle-press} & \textbf{100.0\%} & \textbf{100.0\%} & \textbf{1.00} & \textbf{1.00} \\
        \texttt{coffee-button} & \textbf{85.0\%} & 50.0\% & \textbf{3.70} & 6.90 \\
        \hline
        Overall & \textbf{57.2\%} & 45.0\% & \textbf{5.66} & 6.62 \\
        \hline
    \end{tabular}
    \label{tab:cross_embodiment_context}
\end{table}

\FloatBarrier
\subsection{DIY Unseen-Scene Sweep-Soccer Stress Test}
\label{app:diy_sweep_soccer}

We additionally evaluate VICX on a custom DIY scene outside the standard Meta-World environment family. In this task, the robot must use the gripper to sweep a soccer ball into a hole, introducing a scene layout and object arrangement that do not appear in the benchmark suite used elsewhere in the paper. The goal of this stress test is to assess whether the VICX pipeline can still synthesize useful manipulation plans in a novel scene while keeping the same overall workflow as Appendix~\ref{app:evaluation_results}.

The setting follows the same closed-loop workflow as the rest of the paper. Figure~\ref{fig:diy_sweep_soccer_scene} shows a representative initial scene for this custom task, and Table~\ref{tab:diy_sweep_soccer_context} summarizes the quantitative results. The 0-context setting succeeds on 3 of 20 seeds (15.0\%), while the 5-context setting succeeds on 4 of 20 seeds (20.0\%) and lowers PPC from 9.15 to 8.20. Together, these results show that VICX can transfer non-trivially to a novel DIY scene beyond the standard Meta-World benchmark family, with additional gains from retrieval context.

\begin{center}
\small
\begin{minipage}[t]{0.46\linewidth}
    \vspace{0pt}
    \centering
    \includegraphics[width=0.88\linewidth]{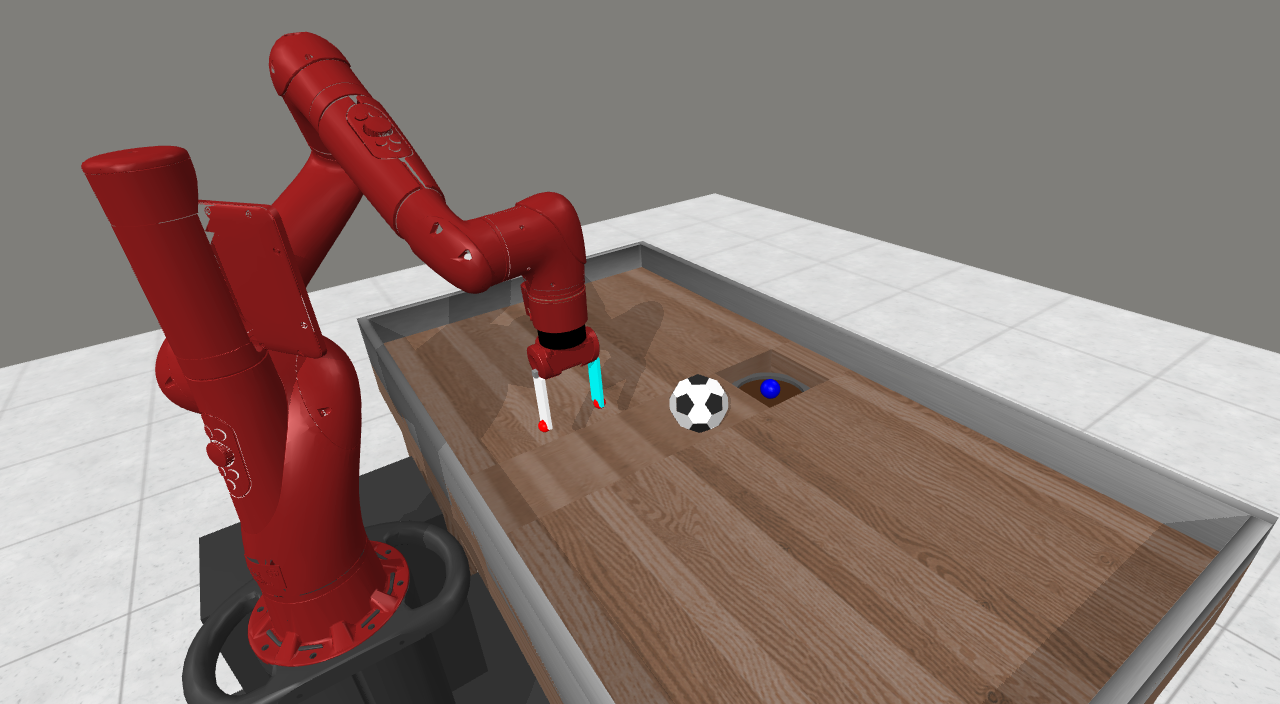}
    \captionof{figure}{Representative initial scene for the custom DIY \texttt{sweep-soccer-into-hole} task.}
    \label{fig:diy_sweep_soccer_scene}
\end{minipage}\hfill
\begin{minipage}[t]{0.50\linewidth}
    \vspace{0pt}
    \centering
    \begin{tabular}{lcc}
        \hline
        Setting & Success Rate $\uparrow$ & PPC $\downarrow$ \\
        \hline
        0 Context & 15.0\% & 9.15 \\
        5 Contexts & \textbf{20.0\%} & \textbf{8.20} \\
        \hline
    \end{tabular}
    \vspace{0.8ex}
    \captionof{table}{Closed-loop evaluation on the custom DIY \texttt{sweep-soccer-into-hole} task. Both settings use the same 20 seeds, planning budget, video model, and V2T-ICON checkpoint. The only difference is whether five retrieved image-state references are provided at inference time. PPC denotes the penalized planning count from Appendix~B.2.}
    \label{tab:diy_sweep_soccer_context}
\end{minipage}
\end{center}

%% file: example.bib
@inproceedings{tian2024robokeygen,
  title={Robokeygen: robot pose and joint angles estimation via diffusion-based {3D} keypoint generation},
  author={Tian, Yang and Zhang, Jiyao and Huang, Guowei and Wang, Bin and Wang, Ping and Pang, Jiangmiao and Dong, Hao},
  booktitle={2024 IEEE International Conference on Robotics and Automation (ICRA)},
  pages={5375--5381},
  year={2024},
  organization={IEEE}
}

@inproceedings{labbe2021single,
  title={Single-view robot pose and joint angle estimation via render \& compare},
  author={Labb{\'e}, Yann and Carpentier, Justin and Aubry, Mathieu and Sivic, Josef},
  booktitle={Proceedings of the IEEE/CVF Conference on Computer Vision and Pattern Recognition},
  pages={1654--1663},
  year={2021}
}

@inproceedings{lee2020camera,
  title={{Camera-to-Robot} pose estimation from a single image},
  author={Lee, Timothy E and Tremblay, Jonathan and To, Thang and Cheng, Jia and Mosier, Terry and Kroemer, Oliver and Fox, Dieter and Birchfield, Stan},
  booktitle={2020 IEEE International Conference on Robotics and Automation (ICRA)},
  pages={9426--9432},
  year={2020},
  organization={IEEE}
}

@article{yang2023context,
  title={In-context operator learning with data prompts for differential equation problems},
  author={Yang, Liu and Liu, Siting and Meng, Tingwei and Osher, Stanley J},
  journal={Proceedings of the National Academy of Sciences},
  volume={120},
  number={39},
  pages={e2310142120},
  year={2023},
  publisher={National Academy of Sciences}
}

@inproceedings{ausserlechner2024zs6d,
  title={{ZS6D}: Zero-shot {6D} object pose estimation using vision transformers},
  author={Ausserlechner, Philipp and Haberger, David and Thalhammer, Stefan and Weibel, Jean-Baptiste and Vincze, Markus},
  booktitle={2024 IEEE International Conference on Robotics and Automation (ICRA)},
  pages={463--469},
  year={2024},
  organization={IEEE}
}

@inproceedings{jantos2023poet,
  title={{PoET}: Pose estimation transformer for single-view, multi-object {6D} pose estimation},
  author={Jantos, Thomas Georg and Hamdad, Mohamed Amin and Granig, Wolfgang and Weiss, Stephan and Steinbrener, Jan},
  booktitle={Conference on Robot Learning},
  pages={1060--1070},
  year={2023},
  organization={PMLR}
}

@inproceedings{
mclean2025metaworld,
title={{Meta-World+}: An Improved, Standardized, {RL} Benchmark},
author={Reginald McLean and Evangelos Chatzaroulas and Luc McCutcheon and Frank R{\"o}der and Tianhe Yu and Zhanpeng He and K.R. Zentner and Ryan Julian and J K Terry and Isaac Woungang and Nariman Farsad and Pablo Samuel Castro},
booktitle={The Thirty-ninth Annual Conference on Neural Information Processing Systems Datasets and Benchmarks Track},
year={2025},
url={https://openreview.net/forum?id=1de3azE606}
}

@article{oquab2024dinov,
title={{DINO}v2: Learning Robust Visual Features without Supervision},
author={Maxime Oquab and Timoth{\'e}e Darcet and Th{\'e}o Moutakanni and Huy V. Vo and Marc Szafraniec and Vasil Khalidov and Pierre Fernandez and Daniel HAZIZA and Francisco Massa and Alaaeldin El-Nouby and Mido Assran and Nicolas Ballas and Wojciech Galuba and Russell Howes and Po-Yao Huang and Shang-Wen Li and Ishan Misra and Michael Rabbat and Vasu Sharma and Gabriel Synnaeve and Hu Xu and Herve Jegou and Julien Mairal and Patrick Labatut and Armand Joulin and Piotr Bojanowski},
journal={Transactions on Machine Learning Research},
issn={2835-8856},
year={2024},
url={https://openreview.net/forum?id=a68SUt6zFt},
note={Featured Certification}
}

@article{su2024roformer,
  title={Roformer: Enhanced transformer with rotary position embedding},
  author={Su, Jianlin and Ahmed, Murtadha and Lu, Yu and Pan, Shengfeng and Bo, Wen and Liu, Yunfeng},
  journal={Neurocomputing},
  volume={568},
  pages={127063},
  year={2024},
  publisher={Elsevier}
}

@InProceedings{black2025pi,
  title = 	 {$\pi_{0.5}$: a Vision-Language-Action Model with Open-World Generalization},
  author =       {Black, Kevin and Brown, Noah and Darpinian, James and Dhabalia, Karan and Driess, Danny and Esmail, Adnan and Equi, Michael Robert and Finn, Chelsea and Fusai, Niccolo and Galliker, Manuel Y. and Ghosh, Dibya and Groom, Lachy and Hausman, Karol and Ichter, Brian and Jakubczak, Szymon and Jones, Tim and Ke, Liyiming and LeBlanc, Devin and Levine, Sergey and Li-Bell, Adrian and Mothukuri, Mohith and Nair, Suraj and Pertsch, Karl and Ren, Allen Z. and Shi, Lucy Xiaoyang and Smith, Laura and Springenberg, Jost Tobias and Stachowicz, Kyle and Tanner, James and Vuong, Quan and Walke, Homer and Walling, Anna and Wang, Haohuan and Yu, Lili and Zhilinsky, Ury},
  booktitle = 	 {Proceedings of The 9th Conference on Robot Learning},
  pages = 	 {17--40},
  year = 	 {2025},
  editor = 	 {Lim, Joseph and Song, Shuran and Park, Hae-Won},
  volume = 	 {305},
  series = 	 {Proceedings of Machine Learning Research},
  month = 	 {27--30 Sep},
  publisher =    {PMLR},
  url = 	 {https://proceedings.mlr.press/v305/black25a.html}
}

@article{wiedemer2025video,
  title={Video models are zero-shot learners and reasoners},
  author={Wiedemer, Thadd{\"a}us and Li, Yuxuan and Vicol, Paul and Gu, Shixiang Shane and Matarese, Nick and Swersky, Kevin and Kim, Been and Jaini, Priyank and Geirhos, Robert},
  journal={arXiv preprint arXiv:2509.20328},
  year={2025}
}

@inproceedings{khachatryan2023text2video,
  title={Text2video-zero: Text-to-image diffusion models are zero-shot video generators},
  author={Khachatryan, Levon and Movsisyan, Andranik and Tadevosyan, Vahram and Henschel, Roberto and Wang, Zhangyang and Navasardyan, Shant and Shi, Humphrey},
  booktitle={Proceedings of the IEEE/CVF International Conference on Computer Vision},
  pages={15954--15964},
  year={2023}
}

@inproceedings{
mirchandani2023large,
title={Large Language Models as General Pattern Machines},
author={Suvir Mirchandani and Fei Xia and Pete Florence and Brian Ichter and Danny Driess and Montserrat Gonzalez Arenas and Kanishka Rao and Dorsa Sadigh and Andy Zeng},
booktitle={7th Annual Conference on Robot Learning},
year={2023},
url={https://openreview.net/forum?id=RcZMI8MSyE}
}

@inproceedings{
dipalo2024kat,
AUTHOR    = {Norman Di Palo AND Edward Johns}, 
    TITLE     = {{Keypoint Action Tokens Enable In-Context Imitation Learning in Robotics}}, 
    BOOKTITLE = {Proceedings of Robotics: Science and Systems}, 
    YEAR      = {2024}, 
    ADDRESS   = {Delft, Netherlands}, 
    MONTH     = {July}, 
    DOI       = {10.15607/RSS.2024.XX.096} 
}

@inproceedings{
vosylius2025instant,
title={Instant Policy: {In-Context} Imitation Learning via Graph Diffusion},
author={Vitalis Vosylius and Edward Johns},
booktitle={The Thirteenth International Conference on Learning Representations},
year={2025},
url={https://openreview.net/forum?id=je3GZissZc}
}

@article{chen2026retrieval,
  title={A retrieval-augmented framework enabling {VLM} spatial awareness for object-centric robot manipulation},
  author={Chen, Kai and Li, Chengkun and Tu, Chang and Pan, Jiahui and Ma, Yiyao and Chen, Wei and Zhou, Zhongxiang and Xu, Xuecheng and James, Stephen and Fu, Chi-Wing and others},
  journal={Science Robotics},
  volume={11},
  number={113},
  pages={eaea2092},
  year={2026},
  publisher={American Association for the Advancement of Science}
}

@article{radosavovic2024real,
  title={Real-world humanoid locomotion with reinforcement learning},
  author={Radosavovic, Ilija and Xiao, Tete and Zhang, Bike and Darrell, Trevor and Malik, Jitendra and Sreenath, Koushil},
  journal={Science Robotics},
  volume={9},
  number={89},
  pages={eadi9579},
  year={2024},
  publisher={American Association for the Advancement of Science}
}

@article{eschmann2026raptor,
  title={RAPTOR: A foundation policy for quadrotor control},
  author={Eschmann, Jonas and Albani, Dario and Loianno, Giuseppe},
  journal={Science Robotics},
  volume={11},
  number={114},
  pages={eaec1481},
  year={2026},
  publisher={American Association for the Advancement of Science}
}

@inproceedings{
liu2025locoformer,
title={{LocoFormer}: Generalist Locomotion via Long-context Adaptation},
author={Min Liu and Deepak Pathak and Ananye Agarwal},
booktitle={9th Annual Conference on Robot Learning},
year={2025},
url={https://openreview.net/forum?id=VqmAvBkFhw}
}

@inproceedings{
fu2024incontext,
title={In-Context Imitation Learning via Next-Token Prediction},
author={Letian Fu and Huang Huang and Gaurav Datta and Lawrence Yunliang Chen and William Chung-Ho Panitch and Fangchen Liu and Hui Li and Ken Goldberg},
booktitle={NeurIPS 2024 Workshop on Open-World Agents},
year={2024},
url={https://openreview.net/forum?id=2R3q4FyPlH}
}

@inproceedings{
sridhar2025ricl,
title={{RICL}:  Adding In-Context Adaptability to Pre-Trained Vision-Language-Action Models},
author={Kaustubh Sridhar and Souradeep Dutta and Dinesh Jayaraman and Insup Lee},
booktitle={9th Annual Conference on Robot Learning},
year={2025},
url={https://openreview.net/forum?id=6AASPlloSt}
}

@inproceedings{
yoo2025robossm,
title={Robo{SSM}: Scalable In-context Imitation Learning via State-Space Models},
author={Youngju Yoo and Jiaheng Hu and Yifeng Zhu and Bo Liu and Qiang Liu and Roberto Mart{\'\i}n-Mart{\'\i}n and Peter Stone},
booktitle={CoRL 2025 Workshop RemembeRL},
year={2025},
url={https://openreview.net/forum?id=JG8p1yGI6U}
}

@inproceedings{finn2017deep,
  title={Deep visual foresight for planning robot motion},
  author={Finn, Chelsea and Levine, Sergey},
  booktitle={2017 IEEE international conference on robotics and automation (ICRA)},
  pages={2786--2793},
  year={2017},
  organization={IEEE}
}

@article{ebert2018visual,
  title={Visual foresight: Model-based deep reinforcement learning for vision-based robotic control},
  author={Ebert, Frederik and Finn, Chelsea and Dasari, Sudeep and Xie, Annie and Lee, Alex and Levine, Sergey},
  journal={arXiv preprint arXiv:1812.00568},
  year={2018}
}

@article{du2023learning,
  title={Learning universal policies via text-guided video generation},
  author={Du, Yilun and Yang, Sherry and Dai, Bo and Dai, Hanjun and Nachum, Ofir and Tenenbaum, Josh and Schuurmans, Dale and Abbeel, Pieter},
  journal={Advances in neural information processing systems},
  volume={36},
  pages={9156--9172},
  year={2023}
}

@inproceedings{du2024video,
  title={Video language planning},
  author={Du, Yilun and Yang, Sherry and Florence, Pete and Xia, Fei and Wahid, Ayzaan and Sermanet, Pierre and Yu, Tianhe and Abbeel, Pieter and Tenenbaum, Joshua B and Kaelbling, Leslie and others},
  booktitle={International Conference on Learning Representations},
  volume={2024},
  pages={31138--31155},
  year={2024}
}

@inproceedings{ko2024learning,
  title={Learning to act from actionless videos through dense correspondences},
  author={Ko, Po-Chen and Mao, Jiayuan and Du, Yilun and Sun, Shao-Hua and Tenenbaum, Joshua B},
  booktitle={International Conference on Learning Representations},
  volume={2024},
  pages={40938--40958},
  year={2024}
}

@inproceedings{wang2025language,
  title={{This$\&$That}: Language-gesture controlled video generation for robot planning},
  author={Wang, Boyang and Sridhar, Nikhil and Feng, Chao and Van der Merwe, Mark and Fishman, Adam and Fazeli, Nima and Park, Jeong Joon},
  booktitle={2025 IEEE International Conference on Robotics and Automation (ICRA)},
  pages={12842--12849},
  year={2025},
  organization={IEEE}
}

@InProceedings{pmlr-v270-liang25b,
  title = 	 {Dreamitate: Real-World Visuomotor Policy Learning via Video Generation},
  author =       {Liang, Junbang and Liu, Ruoshi and Ozguroglu, Ege and Sudhakar, Sruthi and Dave, Achal and Tokmakov, Pavel and Song, Shuran and Vondrick, Carl},
  booktitle = 	 {Proceedings of The 8th Conference on Robot Learning},
  pages = 	 {3943--3960},
  year = 	 {2025},
  volume = 	 {270},
  month = 	 {06--09 Nov},
}

@article{chen2025large,
  title={Large video planner enables generalizable robot control},
  author={Chen, Boyuan and Zhang, Tianyuan and Geng, Haoran and Zhang, Caiyi and Li, Peihao and Song, Kiwhan and Freeman, William T and Malik, Jitendra and Abbeel, Pieter and Tedrake, Russ and others},
  journal={arXiv preprint arXiv:2512.15840},
  year={2025}
}

@article{wang2026world,
  title={World Action Models: The Next Frontier in Embodied {AI}},
  author={Wang, Siyin and Shi, Junhao and Fu, Zhaoyang and He, Xinzhe and Liu, Feihong and Yang, Chenchen and Zhou, Yikang and Fei, Zhaoye and Gong, Jingjing and Fu, Jinlan and others},
  journal={arXiv preprint arXiv:2605.12090},
  year={2026}
}

@article{wan2025wan,
  title={Wan: Open and advanced large-scale video generative models},
  author={{Team Wan} and Wang, Ang and Ai, Baole and Wen, Bin and Mao, Chaojie and Xie, Chen-Wei and Chen, Di and Yu, Feiwu and Zhao, Haiming and Yang, Jianxiao and others},
  journal={arXiv preprint arXiv:2503.20314},
  year={2025}
}

@InProceedings{zitkovich2023rt2,
  title = 	 {{RT-2}: {Vision-Language-Action} Models Transfer Web Knowledge to Robotic Control},
  author =       {Zitkovich, Brianna and Yu, Tianhe and Xu, Sichun and Xu, Peng and Xiao, Ted and Xia, Fei and Wu, Jialin and Wohlhart, Paul and Welker, Stefan and Wahid, Ayzaan and Vuong, Quan and Vanhoucke, Vincent and Tran, Huong and Soricut, Radu and Singh, Anikait and Singh, Jaspiar and Sermanet, Pierre and Sanketi, Pannag R. and Salazar, Grecia and Ryoo, Michael S. and Reymann, Krista and Rao, Kanishka and Pertsch, Karl and Mordatch, Igor and Michalewski, Henryk and Lu, Yao and Levine, Sergey and Lee, Lisa and Lee, Tsang-Wei Edward and Leal, Isabel and Kuang, Yuheng and Kalashnikov, Dmitry and Julian, Ryan and Joshi, Nikhil J. and Irpan, Alex and Ichter, Brian and Hsu, Jasmine and Herzog, Alexander and Hausman, Karol and Gopalakrishnan, Keerthana and Fu, Chuyuan and Florence, Pete and Finn, Chelsea and Dubey, Kumar Avinava and Driess, Danny and Ding, Tianli and Choromanski, Krzysztof Marcin and Chen, Xi and Chebotar, Yevgen and Carbajal, Justice and Brown, Noah and Brohan, Anthony and Arenas, Montserrat Gonzalez and Han, Kehang},
  booktitle = 	 {Proceedings of The 7th Conference on Robot Learning},
  pages = 	 {2165--2183},
  year = 	 {2023},
  volume = 	 {229},
  publisher =    {PMLR},
}

@inproceedings{kim2025openvla,
  title = {{OpenVLA}: An Open-Source Vision-Language-Action Model},
  author = {Kim, Moo Jin and Pertsch, Karl and Karamcheti, Siddharth and Xiao, Ted and Balakrishna, Ashwin and Nair, Suraj and Rafailov, Rafael and Foster, Ethan P. and Sanketi, Pannag R. and Vuong, Quan and Kollar, Thomas and Burchfiel, Benjamin and Tedrake, Russ and Sadigh, Dorsa and Levine, Sergey and Liang, Percy and Finn, Chelsea},
  booktitle = {Proceedings of The 8th Conference on Robot Learning},
  series = {Proceedings of Machine Learning Research},
  volume = {270},
  pages = {2679--2713},
  year = {2025}
}

@inproceedings{oneill2024openx,
  title = {{Open X-Embodiment}: Robotic Learning Datasets and {RT-X} Models},
  author={O'Neill, Abby and Rehman, Abdul and Maddukuri, Abhiram and Gupta, Abhishek and Padalkar, Abhishek and Lee, Abraham and Pooley, Acorn and Gupta, Agrim and Mandlekar, Ajay and Jain, Ajinkya and others},
  booktitle={2024 IEEE International Conference on Robotics and Automation (ICRA)},
  pages={6892--6903},
  year={2024},
}

@ARTICLE{ma2026surveyofvla,
  author={Ma, Yueen and Song, Zixing and Zhuang, Yuzheng and Hao, Jianye and King, Irwin},
  journal={IEEE Transactions on Neural Networks and Learning Systems}, 
  title={A Survey on Vision–Language–Action Models for Embodied {AI}}, 
  year={2026},
  pages={1-21},
  doi={10.1109/TNNLS.2025.3650584}}

@article{douze2026faiss,
  title={The {Faiss} library},
  author={Douze, Matthijs and Guzhva, Alexandr and Deng, Chengqi and Johnson, Jeff and Szilvasy, Gergely and Mazar{\'e}, Pierre-Emmanuel and Lomeli, Maria and Hosseini, Lucas and J{\'e}gou, Herv{\'e}},
  journal={IEEE Transactions on Big Data},
  year={2026},
  doi={10.1109/TBDATA.2025.3618474}
}

@inproceedings{
ye2026world,
title={World Action Models are Zero-shot Policies},
author={Seonghyeon Ye and Yunhao Ge and Kaiyuan Zheng and Shenyuan Gao and Sihyun Yu and George Kurian and Suneel Indupuru and You Liang Tan and Chuning Zhu and Jiannan Xiang and Ayaan Naveed Malik and Kyungmin Lee and William Liang and Nadun Ranawaka Arachchige and Jiasheng Gu and Yinzhen Xu and Guanzhi Wang and Fengyuan Hu and Avnish Narayan and Johan Bjorck and Jing Wang and Gwanghyun Kim and Dantong Niu and Ruijie Zheng and Yuqi Xie and Jimmy Wu and Qi Wang and Danfei Xu and Yilun Du and Ryan Julian and Yevgen Chebotar and Scott Reed and Jan Kautz and Yuke Zhu and Linxi Fan and Joel Jang},
booktitle={ICLR 2026 the 2nd Workshop on World Models: Understanding, Modelling and Scaling},
year={2026},
url={https://openreview.net/forum?id=cd33uUB609}
}
